\documentclass[10pt,journal,compsoc]{IEEEtran}
\ifCLASSOPTIONcompsoc
  \usepackage[nocompress]{cite}
  \usepackage[colorlinks=false,linkcolor=black,citecolor=black]{hyperref}
  \hypersetup{hidelinks} 
  \usepackage[utf8]{inputenc}
  \usepackage{graphicx}
  \usepackage{caption}
  \usepackage{subfigure} 
  \usepackage{multirow} 
  \usepackage{makecell} 
  \usepackage{float}
  \usepackage{amsmath} 
  \usepackage{amsfonts}
  \usepackage{booktabs}
  \usepackage{color}
  \usepackage{newtxmath}

  \usepackage{bbding}
  \usepackage{pifont}
  \usepackage{wasysym}
\else
  \usepackage{cite}
\fi
\ifCLASSINFOpdf
\else
\fi
\newcommand{\tabincell}[2]{\begin{tabular}{@{}#1@{}}#2\end{tabular}}

\begin{document}
\title{FGAHOI: Fine-Grained Anchors for\\Human-Object Interaction Detection}

\author{Shuailei~Ma,~\IEEEmembership{}
        Yuefeng~Wang,~\IEEEmembership{}
        Shanze~Wang,~\IEEEmembership{}
        and~Ying~Wei~\IEEEmembership{}% <-this % stops a space
\IEEEcompsocitemizethanks{\IEEEcompsocthanksitem Shuailei Ma, Yuefeng Wang are with College of Information Science and Engineering, Northeastern University, Shenyang, China, 110819.\protect\\
E-mail: \{xiaomabufei, wangyuefeng0203\} @gmail.com

\IEEEcompsocthanksitem Shanze Wang is with Changsha Hisense Intelligent System Research Institute Co., Ltd. and Information Technology R\&D Innovation Center of Peking University, Shaoxing, China.\\
E-mail: szgg0099@gmail.com

\IEEEcompsocthanksitem Ying Wei is the corresponding author, with College of Information Science and Engineering, Northeastern University, Shenyang, China, 110819.\\
E-mail: weiying@ise.neu.edu.cn
}% <-this % stops a space

\thanks{Manuscript received October 26, 2022; revised January 10, 2023.}}


\markboth{Journal of \LaTeX\ Class Files,~Vol.~14, No.~10, January~2023}%
{Shell \MakeLowercase{\textit{et al.}}: Bare Advanced Demo of IEEEtran.cls for IEEE Computer Society Journals}



\IEEEtitleabstractindextext{%
\begin{abstract}
Human-Object Interaction (HOI), as an important problem in computer vision, requires locating the human-object pair and identifying the interactive relationships between them. The HOI instance has a greater span in spatial, scale, and task than the individual object instance, making its detection more susceptible to noisy backgrounds. To alleviate the disturbance of noisy backgrounds on HOI detection, it is necessary to consider the input image information to generate fine-grained anchors which are then leveraged to guide the detection of HOI instances. However, it is challenging for the following reasons. $i)$ how to extract pivotal features from the images with complex background information is still an open question. $ii)$ how to semantically align the extracted features and query embeddings is also a difficult issue. In this paper, a novel end-to-end transformer-based framework (FGAHOI) is proposed to alleviate the above problems. FGAHOI comprises three dedicated components namely, \textbf{multi-scale sampling (MSS)}, \textbf{hierarchical spatial-aware merging (HSAM)} and \textbf{task-aware merging mechanism (TAM)}. MSS extracts features of humans, objects and interaction areas from noisy backgrounds for HOI instances of various scales. HSAM and TAM semantically align and merge the extracted features and query embeddings in the hierarchical spatial and task perspectives in turn. In the meanwhile, a novel training strategy \textbf{Stage-wise Training Strategy} is designed to reduce the training pressure caused by overly complex tasks done by FGAHOI. In addition, we propose two ways to measure the difficulty of HOI detection and a novel dataset, $i.e.$, HOI-SDC for the two challenges (\textbf{Uneven Distributed Area in Human-Object Pairs} and \textbf{Long Distance Visual Modeling of Human-Object Pairs}) of HOI instances detection. Experiments are conducted on three benchmarks: HICO-DET, HOI-SDC and V-COCO. Our model outperforms the state-of-the-art HOI detection methods, and the extensive ablations reveal the merits of our proposed contribution. The code is available at \url{https://github.com/xiaomabufei/FGAHOI}.
% However, this is challenging due to extract pivotal features from the images with complex background information is still an open question. Besides, how to semantically align the extracted features and query embeddings is also a difficult issue. 

\end{abstract}

}

\maketitle

% is not selected <OR> if conference mode is selected - because compsoc
% conference papers position the abstract like regular (non-compsoc)
\IEEEdisplaynontitleabstractindextext

% \fi
\IEEEpeerreviewmaketitle

\ifCLASSOPTIONcompsoc
\IEEEraisesectionheading{\section{Introduction}\label{sec:introduction}}
\else
\section{Introduction}
\label{sec:introduction}
\fi

\IEEEPARstart{H}{UMAN}-Object interaction (HOI) detection, as a downstream task of object detection \cite{r16,r22,r31,r40,r46}, has recently received increasing attention due to its great application potential. For successful HOI detection, it needs to have the ability to understand human activities which are abstracted as a set of $<$human, object, action$>$ triplets in this task, requiring a much deeper understanding for the semantic information of visual scenes. Without HOI detection, machines can only interpret images as collections of object bounding boxes, i.e., AI systems can only pick up information such as 'A man is on the bike' or 'A bike is in the corner', but not 'A man rides a bike'.
% \begin{figure}[htbp]
%     \centering
%     \includegraphics[width = \linewidth]{figure22.pdf}
%     \setlength{\abovecaptionskip}{-0.2cm}
%     \setlength{\belowcaptionskip}{-0.5cm}
%     \caption{Illustration of the difference between object and HOI detection. (a) Object detection only requires detecting the object instance (contain human instance). (b) HOI detection requires not only to detect the human-object pair which consists of one human and object, but also to identify the interaction relationship (action class).}
%     \label{fig:Figure1}
% \end{figure}
\begin{figure}[htbp]
    \centering
    \includegraphics[width = \linewidth]{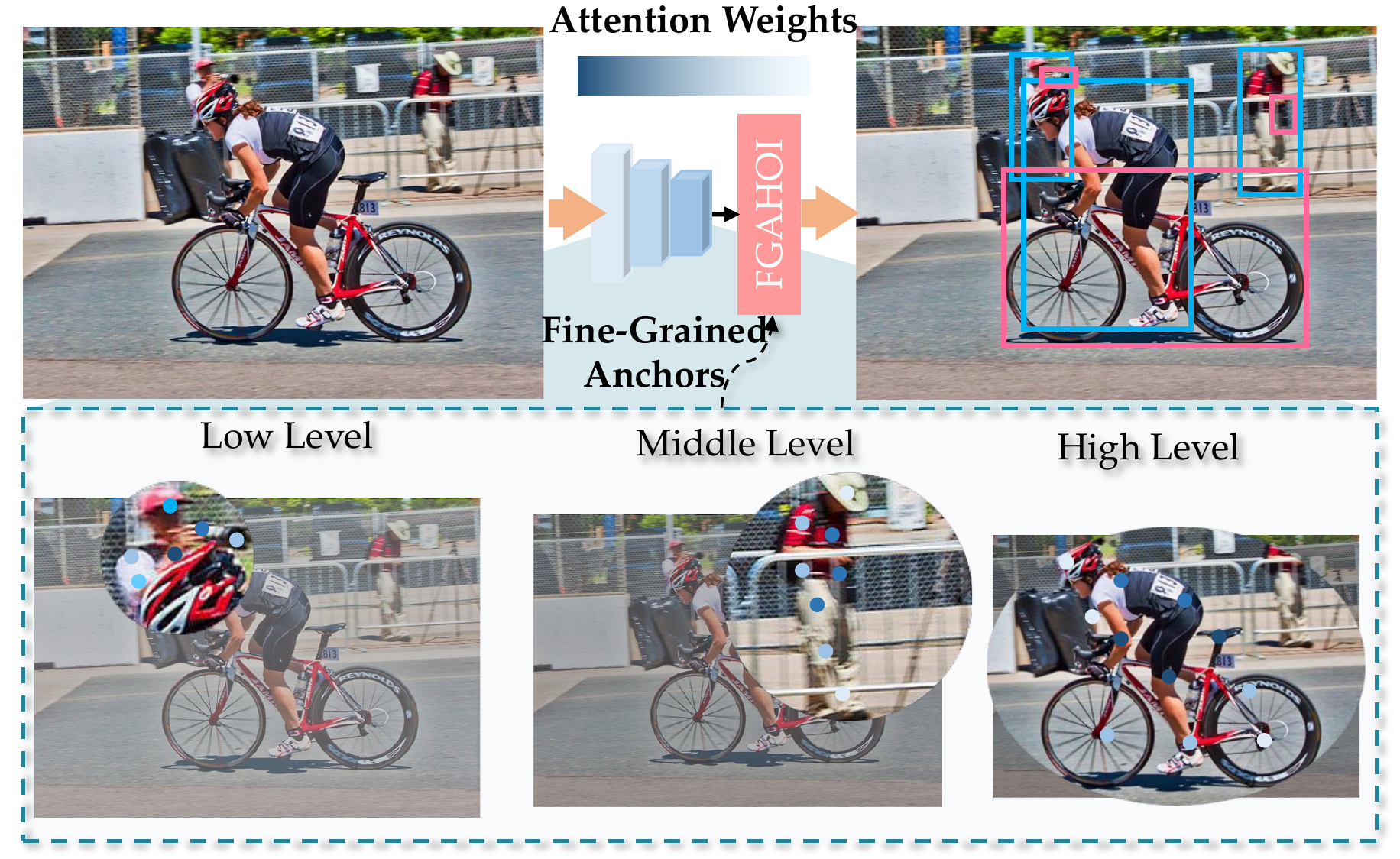}
    \setlength{\abovecaptionskip}{-0.2cm}
    \setlength{\belowcaptionskip}{-0.5cm}
    \caption{FGAHOI leverages the query embeddings and multi-scale features to generate fine-grained anchors and the corresponding weights for HOI instances of diverse scales. Then, they guide the decoder to aid key semantic information of HOI instances to the content embeddings and translate the content embeddings to HOI embeddings for predicting all elements of the HOI instances.}
    \label{fig:Figure1}
\end{figure}

Spanning the past and the present, the existing HOI detection approaches \cite{r2,r4,r5,r7,r8,r9,r10,r14,r17,r19,r20,r21,r24,r33,r39,r47} tend to fall into two categories, namely two-stage and one-stage methods. Conventional two-stage methods \cite{r4,r5,r8,r10,r14,r17,r24,r30,r38,r39,r43,r44}, as an intuitive approach, detect human and object instances by leveraging the off-the-shelf object detector \cite{r16,r31,r40}, utilizing the visual features extracted from the located areas to recognize action classes. To fully leverage the visual features, several methods \cite{r4,r8,r17,r30,r38,r39,r43,r44} separately extract visual features of human-object pairs and spatial information from the located area in a multi-stream architecture, fusing them in a post-fusion strategy. In the meanwhile, several approaches \cite{r5,r8,r38,r39,r43} employ the existing pose estimation methods, such as \cite{r12,r32,r37} to extract pose information and fuse it with other features to predict the action class. In addition, some works \cite{r5,r10,r14,r24,r29} leverage the graph neural network to extract complex semantic relationship between humans and objects. However, the difficulties encountered in the two-stage approach lie mainly in the effective fusion of human-object pairs and complex semantic information. Besides, owing to the limitations of the fixed detector and some other components (pose estimation etc.), the two-stage method can only achieve a sub-optimal solution. 

To achieve high efficiency, one-stage approaches \cite{r2,r7,r9,r19,r21,r41,r45,r47} which utilize interaction points between the human-object pairs to simultaneously predict human and object offset vectors and action classes, are proposed to detect human-object pairs and recognize interactive relationships in parallel. However, when the human and object in the image are far apart from each other, these methods are disturbed by ambiguous semantic features. The one-stage methods do not achieve much attention until the appearance of the Detection Transformer (DETR) \cite{r3} and QPIC \cite{r33} applies it for HOI detection. Then, plenty of transformer-based works \cite{r20, r34, r21, r2, r7} attempt to solve the HOI detection with different encoder-decoder structures and backbone models.

In comparison to object instances, HOI instances have a greater span of spatial, scale and task. In most HOI instances, there is a certain distance between human and objects and their scale varies enormously. Compared with simple object classification, it is necessary to consider more information between human-object pairs rather than the features of humans and objects for interaction classification. Therefore, the detection is more susceptible to distractions from noisy backgrounds. However, most recent works \cite{r33, r34} use object detection frameworks \cite{r3, r15} directly for HOI detection by simply adding the interaction classification head, ignoring these problems. Inspired by \cite{r15} which leverages the reference points to guide the decoding process, we propose to leverage fine-grained anchors to guide the detection of HOI instances and protect it from noisy backgrounds. To generate fine-grained anchors for kinds of HOI instances, it is obviously necessary to consider the input image features. There are, however, two inevitable challenges that arise as a result of this. $i)$ it is difficult to extract pivotal features from the images which contain noisy background information. $ii)$ how to semantically align and merge the extracted features with query embeddings is also an open question.

In this paper, we propose a novel transformer-based model for HOI detection, i.e., FGAHOI: Fine-Grained Anchors for Human-Object Interaction Detection (as shown in Fig.\ref{fig:Figure1}). FGAHOI leverages the \textbf{multi-scale sampling mechanism (MSS)} to extract pivotal features from images with noisy background information for variable HOI instances. Based on the sampling strategy and initial anchor generated by the corresponding query embedding, MSS could extract hierarchical spatial features of human, object and the interaction region for each HOI instance. Besides, the \textbf{hierarchical spatial-aware (HSAM)} and \textbf{task-aware merging mechanism (TAM)} are utilized to semantically align and merge the extracted features with the query embeddings. HSAM merges the extracted features in the hierarchical spatial perspective according to the cross-attention between the features and the query embeddings. Meanwhile, the extracted features are aligned towards the query embeddings, according to the cross-attention weights of the merging process. Thereafter, TAM leverages the switches which dynamically switch ON and OFF to merge the input features and query embeddings in the task perspective.

According to experiment results, we investigate that it is difficult of the end-to-end training approach to allow the transformer-based models to achieve optimal performance when more complex task requirements are required. Inspired by the stage-wise training \cite{r73,r74} for LTR \cite{r75}, we propose a novel stage-wise training strategy for FGAHOI. During the training process, we add the important components of the model in turn to clarify the training direction of the model at each stage, so as to maximize the savings in the training cost of the model.

To the best of our knowledge, there are no measurements for the difficulty of detecting HOI instances. We investigate that two difficulties lie in the detection of human-object pairs, $i.e.$, \textbf{Uneven Distributed Area in Human-Object Pairs} and \textbf{Long Distance Visual Modeling of Human-Object Pairs}. In this paper, we propose two measurements and a novel dataset (HOI-SDC) for these two challenges. HOI-SDC eliminates the influence of other factors (Too few training samples of some HOI categories, too tricky interaction actions, et.al.) on the model training and focuses on the model for these two difficult challenges.
Our contributions can be summarized fourfold: 
\begin{itemize}
\item[$\bullet$]We propose a novel transformer-based human-object interaction detector (FGAHOI) which leverages input features to generate fine-grained anchors for protecting the detection of HOI instances from noisy backgrounds.
\item[$\bullet$]We propose a novel training strategy where each component of the model is trained in turn to clarify the training direction at each stage, in order to maximize the training cost savings.
\item[$\bullet$]We propose two ways to measure the difficulty of HOI detection and a dataset, $i.e.$, HOI-SDC for the two challenges (Uneven Distributed Area in Human-Object Pairs and Long Distance Visual Modeling of Human-Object Pairs) of detecting HOI instances.
\item[$\bullet$]Our extensive experiments on three benchmarks: HICO-DET \cite{r54}, HOI-SDC and V-COCO \cite{r59}, demonstrate the effectiveness of the proposed FGAHOI. Specifically, FGAHOI outperforms all existing state-of-the-art methods by a large margin.
\end{itemize}
\begin{figure*}[htbp]
    \centering
    \includegraphics[width = \textwidth]{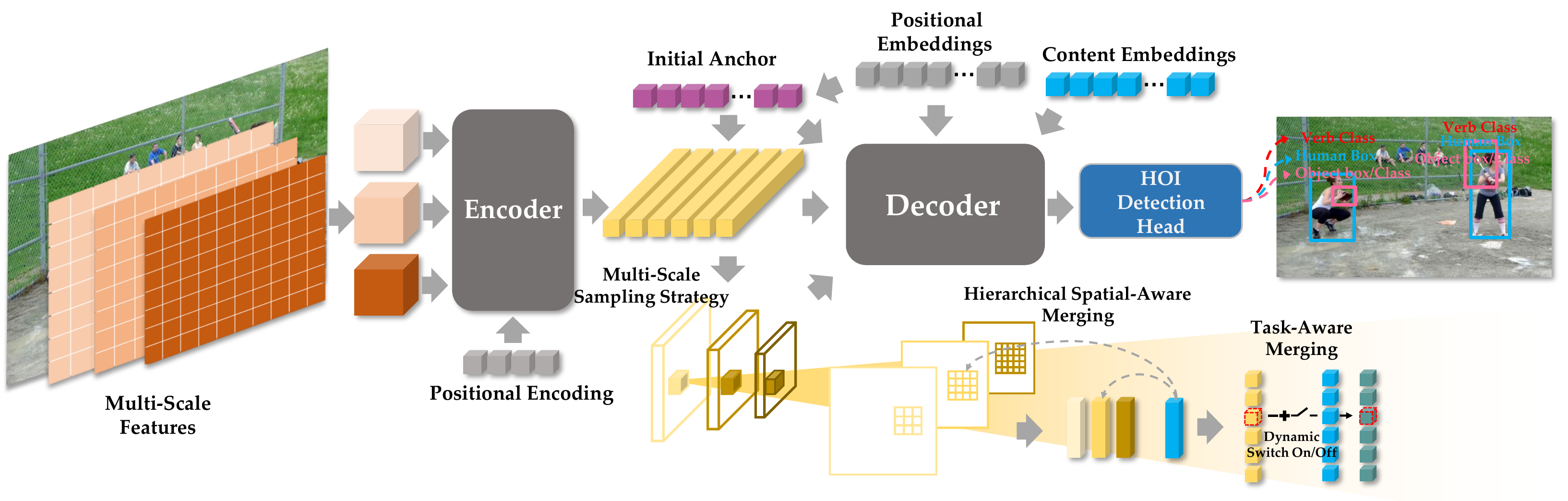}
    \caption{This figure illustrates the overall structure of FGAHOI. FGAHOI utilizes a hierarchical backbone and a deformable encoder to extract the semantic features in a multi-scale approach. In the decoding phrase, FGAHOI leverages the \textbf{multi-scale sampling}, \textbf{hierarchical spatial-aware merging} and \textbf{task-aware merging} mechanism to align input features with query embeddings and assist the generation of fine-grained anchors for the translation of HOI embeddings. At the back end of the pipeline, HOI detector leverages the HOI embeddings and initial anchor to predict all elements of the HOI instances.}
    \label{fig:Figure2}
\end{figure*}
\section{Related Works}
\textbf{Two-stage HOI Detection Approaches}: 
The two-stage HOI detection approaches \cite{r4,r5,r8,r10,r14,r17,r24,r29,r30,r38,r39,r43,r44} employ the off-the-shelf object detector \cite{r16,r31,r40} to localize humans and objects. Afterwards, the features of backbone networks inside the human and objects regions are cropped. Part of the two-stage methods \cite{r5,r10,r14,r24,r29} treat the human and objects feature as nodes and employ graph neural networks \cite{r48} to predict action classes. The other part of the two-stage approach \cite{r4,r8,r17,r30,r38,r39,r43,r44} leverages multi-stream networks to extract diverse information from cropped regions, such as human features, object features, spatial information and human pose information. Then, the information is fused to predict the action in a post-fusion strategy. Two-stage methods mainly concentrate on predicting the action class in the second stage. Nevertheless, the quality of cropped features from the first stage cannot be guaranteed in most cases, so the method cannot achieve an optimal solution. More importantly, integrating semantic information of human-object pairs requires massive time and computing resources.\par

\noindent \textbf{One-stage HOI Detection Approaches:}
The traditional one-stage approaches \cite{r7,r9,r19,r45} use interaction points or union regions to detect human-object pairs and identify interactive action classes in parallel. However, these methods which and are hampered by distant human-object pairs,  require a gathering and pairing process. With the creation of DETR \cite{r3}, one-stage approaches have become the current mainstream. QPIC \cite{r33} converts the object detection head of DETR into an interaction detection head to predict HOI instance directly. HOITrans \cite{r21} combines transformer \cite{r11} and CNN \cite{r49} to straightly predict HOI instances from the query embeddings. AS-Net \cite{r2} and HOTR \cite{r7} each propose a two-branch transformer method that consists of an instance decoder and an interaction decoder to predict the boxes and action classes in parallel. CDN \cite{r20} proposes a cascade disentangling decoder to decode action classes. QAHOI \cite{r34} directly combines Swin Transformer \cite{r27} and deformable DETR \cite{r15} to predict HOI instances. \par
\noindent \textbf{Anchor-Based Object Detection Transformer:} Deformable DETR \cite{r15} first introduces the reference point concept, where the sampling offset is predicted by each reference point to perform deformable cross-attention. To facilitate extreme region discrimination, Conditional DETR \cite{r63} reformulates the attention operation and rebuilt positional queries based on reference points. Anchor DETR \cite{r60} proposes to explicitly capitalize on the spatial prior during cross-attention and box regression by utilizing a predefined 2D anchor point $[cx,cy]$. DAB-DETR \cite{r61} extends such a 2D concept to a 4D anchor box $[cx,cy,w,h]$ and proposed to refine it layer-by-layer. SAM-DETR \cite{r64} proposes directly updating content embeddings by extracting salient points from image features. In this paper, we propose a novel decoding process for HOI detection. The alignment and fine-grained anchor generation is proposed to align the multi-scale features with HOI query embeddings and generate fine-grained anchors for the diverse HOI instances with variable spatial distribution, scales and tasks. Then, the fine-grained anchors guide the deformable attention process in aiding key information to query embeddings from noisy backgrounds.

\section{Proposed Method}
In Sec.\ref{sec.1}, we show the overall architecture of FGAHOI. Then, we describe the multi-scale feature extractor in Sec.\ref{sec.2}. We introduce the multi-scale sampling strategy in Sec.\ref{sec.4}. The hierarchical spatial-aware, task-aware merging mechanism and the decoding process is proposed in Sec.\ref{sec.5}, Sec.\ref{sec.7} and Sec.\ref{sec.8}, respectively. In Sec.\ref{sec.9}, we present the architecture of the HOI detection head. In Sec.\ref{sec.10}, the stage-wise training strategy, loss calculation and inference process is illustrated.
\subsection{Overall Architecture}\label{sec.1}
The overall architecture of our proposed FGAHOI is illustrated in Fig \ref{fig:Figure2}. For a given image ${x} \in \mathbb{R}^{{H}\times{W}\times{3}}$, FGAHOI firstly uses a hierarchical backbone network to extract the multi-scale features $\mathrm{Z}_{i} \in \mathbb{R}^{\frac{H}{4 \times 2^{i}} \times \frac{W}{4 \times 2^{i}} \times 2^{i} C_{s}},i=1,2,3 $. The multi-scale features are then projected from dimension $\mathrm{C}_{s}$ to dimension $\mathrm{C}_{d}$ by using 1×1 convolution. After being flattened out, the multi-scale features are concatenated to $\mathrm{N}_{s}$ vectors with $\mathrm{C}_{d}$ dimensions. Afterwards, along with supplementary positional encoding $p \in \mathbb{R}^{N_{s} \times C_{d}}$, the multi-scale features are sent into the deformable transformer encoder which consists of a set of stacked deformable encoder layers to encode semantic features. The encoded semantic features $M \in \mathbb{R}^{N_{s} \times C_{d}}$ are then acquired. In the decoding process, the content $C$ and positional $P$ embeddings are both a set of learnable vectors $\left\{v_{i} \mid v_{i} \in \mathbb{R}^{c_{d}}\right\}_{i=1}^{N_{q}}$. The positional embeddings $P$ first generate the initial anchor $A \in \mathbb{R}^{N_q \times 2}$ according to a linear layer. The positional $P$, content $C$ embeddings, inital anchor $A$ and encoded features $M$ are simultaneously sent into the decoder $F_{decoder}(\cdot,\cdot,\cdot,\cdot)$ which is a set of stacked decoder layers. In every decoder layer, the initial anchor first leverages the multi-scale sampling strategy to sample the multi-scale features corresponding to the content embeddings. The sampled features assist the generation of fine-grained anchors and corresponding attention weights through the hierarchical spatial-aware and task-aware merging mechanism. The HOI embeddings $H=\left\{h_{i} \mid h_{i} \in  \mathbb{R}^{c_{d}}\right\}_{i=1}^{N_{q}}$ are translated from the query embeddings $Q$ through the fine-grained anchors, attention weights and the deformable attention. The HOI embeddings $H$ are acquired as $H=F_{decoder}(M, P, C, A)$. Eventually, the HOI detector leverages the HOI embeddings $H$ and initial anchor to predict the HOI instances $<b_h, b_o, c_o, c_v>$, where $b_h$, $b_o$, $c_o$ and $c_v$ stands for the human box coordinate ($x,y,w,h$), object box coordinate, object class and verb class, respectively.
\begin{figure*}[htbp]
    \centering
    \includegraphics[width = \textwidth]{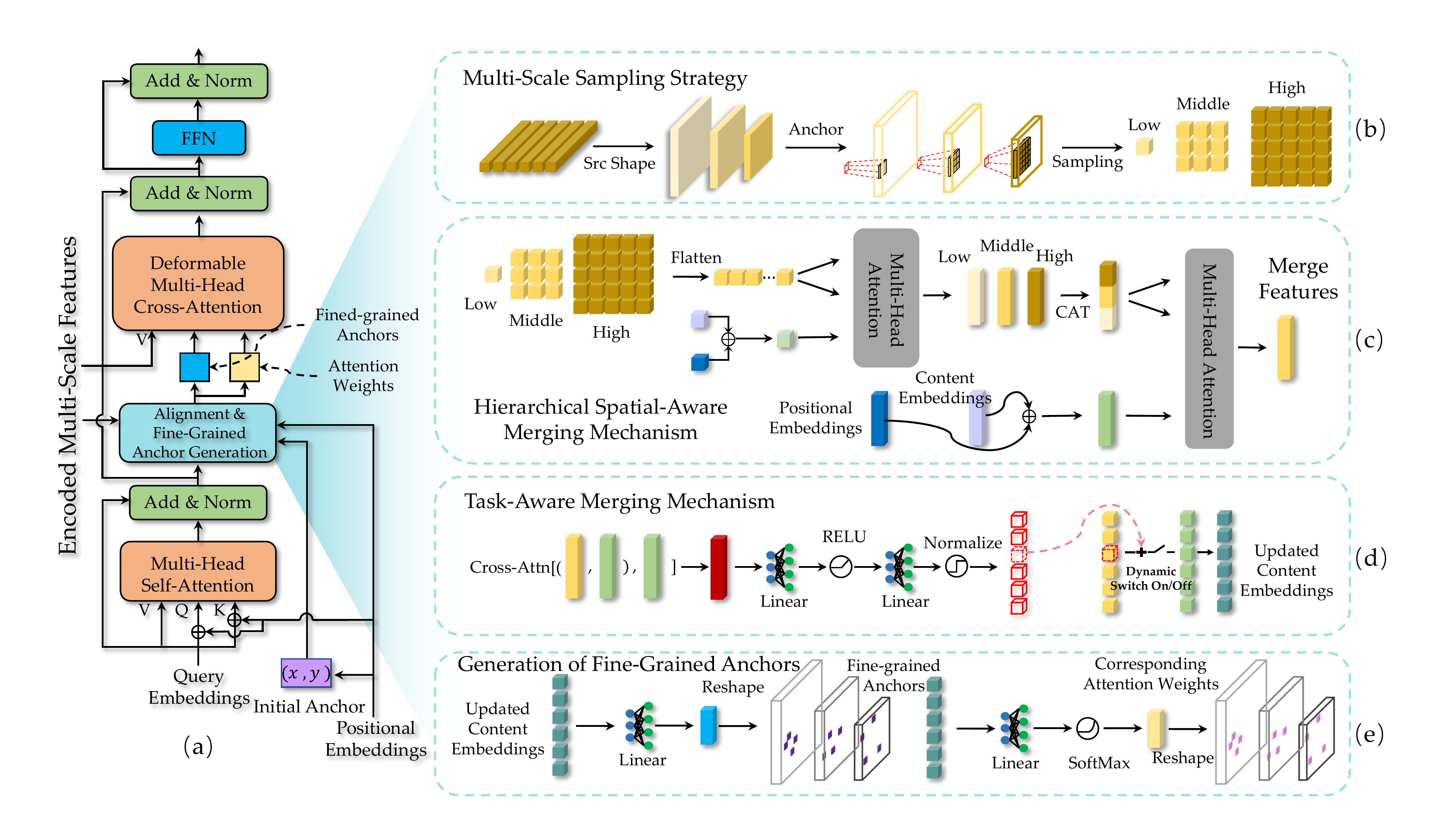}
    \caption{The architecture of FGAHOI's decoder. (a) Illustration of FGAHOI's decoding process. (b) Illustration of Multi-scale sampling mechanism. (c) Illustration of Hierarchical spatial-aware merging mechanism. (d) Illustration of Task-aware merging mechanism. (e) Generation process of fine-grained anchors and the corresponding attention weights.}
    \label{fig:Figure3}
\end{figure*}
\subsection{Multi-Scale Features Extractor}\label{sec.2}
High-quality visual features are a prerequisite for successful HOI detection.
For extracting the multi-scale features with long-range semantic information, FGAHOI leverages the multi-scale feature extractor which consists of a hierarchical backbone network and a deformable transformer encoder to extract features, the folumation is as Equation.\ref{eq1}:
\begin{equation}\label{eq1}
M=F_{encoder}(F_{flatten}(\phi(x)), p, s, r, l)\ \ \in \mathbb{R}^{N_s \times C_d},
\end{equation}
where $F_{encoder}(\cdot)$, $F_{flatten}(\cdot)$ and $\phi(\cdot)$ denotes the encoder, flatten operation and backbone network, respectively. $p$ is the position encoding, $s$ is the spatial shape of the multi-scale features, $r$ stands for the  valid ratios and $l$ represents the level index corresponding the multi-scale features. The hierarchical backbone network is flexible and can be composed of any convolutional neural network \cite{r49,r70,r71,r72} and transformer backbone network \cite{r1,r13,r23,r26,r27,r35,r36,r42}. However, CNN is poor at capturing non-local semantic features like the relationships between humans and objects. In this paper, we mainly use Swin Transformer tiny and large version \cite{r27} to enhance the ability of feature extractor for extracting long-range features.

\subsection{Why FGAHOI Decodes Better?}\label{sec.3}
During the decoding process, the fine-grained anchors can be regarded as a positional prior to let decoder focus on the region of interest, directly guiding the decoder to aid semantic information to the content embeddings which are used to predict all elements of the HOI instances. Therefore, fine-grained anchors play the following two crucial roles in HOI detection. $i)$ Fine-grained anchors directly determine whether the information gained from input features to content embeddings is instance-critical or noisy background information. $ii)$ Fine-grained anchors determine the quality of alignment between the query embeddings and multi-scale features of input scenarios. Both are crucial factors for the quality of decoding results. The existing methods \cite{r15,r34} directly utilize the query embedding to generate fine-grained anchors based on the initial anchor, without considering the multi-scale features of the input scenarios and the semantic alignment between the query embedding and the input features at all. Our FGAHOI proposes a novel fine-grained anchors generator which consists of \textbf{multi-scale sampling}, \textbf{hierarchical spatial-aware merging} and \textbf{task-aware merging mechanism} (as shown in Fig.\ref{fig:Figure3}). The generator adequately leverages the initial anchor, multi-scale features and query embeddings for generating suitable fine-grained anchors for diverse input scenarios and aligning semantic information between different input scenarios and query embeddings. The formulation of FGAHOI decoding process is as follows:
\begin{equation}
\begin{aligned}
H=\operatorname{Defattn}(\operatorname{Task}(\operatorname{Hier\ Spatial}(\{x_{s}^i\},C_u),C_u),M,C_u),
\end{aligned}
\end{equation}
where $C_u$ is the content embeddings updated by the positional embeddings, $\operatorname{Defattn}$ represents the deformable attention, $x_{s}^i$ represents the sampled features of the $i$-th level features. $M$ is the encoded input features.

\subsubsection{Multi-Scale Sampling Mechanism}\label{sec.4}
The HOI instances contained in the input scenarios usually vary in size, where some instances taking up most of the area in the input scenarios and others occupying perhaps only a few pixels. Our FGAHOI aims at detecting all instances in the scene, regardless of the size. Therefore, when using the initial anchor to sample the multi-scale features, for shallow features mainly used to detect instances of small size, the sampling strategy only samples a small range of features around the initial anchor. In contrast, for deep features mainly used to detect instances of large size, the sampling strategy samples a large range of features around the initial anchor.
As shown in Fig.\ref{fig:Figure3} (b), in the generator, the encoded features are first reshaped to the original shape. Based on the initial anchor, generator leverages the sampling strategy to sample multi-scale features as follows:
\begin{equation}
\begin{aligned}
\centering
x_{s}^i=& F_{sample}(\ reshape(M)^i, A,{size}^i,bilinear\ ), \\
% & \ \ \ \ \ \ \ \ \ \ \ \ \ \ \ \ \ \ \ \ \ \ \ \ \ \ \ \ \ \ 
% \ \ \in \mathbb{R}^{B \times N \times\left({size}^i\right)^2 \times C}
\end{aligned}
\end{equation}
where ${size}^i$ ($i=0,1,2$) denotes the sampling size of the $i$-th level features. $M$ is the encoded input features. $A$ is the initial anchor. Inspired by \cite{r68}, we utilize bilinear interpolation in the sampling strategy.

\subsubsection{Hierarchical Spatial-Aware Merging Mechanism}\label{sec.5}
In order to better utilize the hierarchical spatial information of sampled features for aligning content embeddings with the sampled features, we propose a novel hierarchical spatial-aware merging mechanism (HSAM) which utilizes the content embeddings to extract hierarchical spatial information and merge the sampled features, as shown in Fig.\ref{fig:Figure3} (c). The content embeddings are first updated by the positional embeddings and multi-head self-attention mechanism as follows:
\begin{equation}
\begin{aligned}
\centering
C_u= C + F_{\text{\tiny MHA}}\left((C+P)W^q,(C+P)W^k,CW^v\right), 
% \\
% &\ \ \ \ \ \ \  \ \ \ \ \ \ \ \ \ \ \ \ \ \ \ \ \ \ \ \ \ \ \ \ \ \ \ \ \ 
% \ \ \ \in \mathbb{R}^{B \times N \times C}
\end{aligned}
\end{equation}
where $W^q$, $W^k$ and $W^v$ denotes the parameter matrices for query, key and value in the self-attention mechanism, respectively. $F_{\text{\tiny MHA}}(\cdot)$ is the multi-head attention mechanism. $C$ and $P$ represents the content and position embeddings, respectively. Then, the updated content embeddings are leveraged to merge the sampled features, the formulation is as follows:
% \begin{equation}
% \begin{aligned}
% \centering
% x_{m}^i= F_{\text{\tiny MHA}}\left(C_u W^q,x_{s}^iW^k,x_{s}^iW^v\right),
% % \in \mathbb{R}^{B \times N \times C}
% \end{aligned}
% \end{equation}
\begin{equation}
\begin{aligned}
x_{m}^i &=F_{\operatorname{concat}}\left(\operatorname{head}_1, \ldots, \operatorname{head}_{\mathrm{N_H}}\right) W^O, \\
 \operatorname{ where }\ \ {\operatorname{head}}_{\operatorname{n}} &=\operatorname{Softmax} \left(\frac{(C_u W^q_{\operatorname{n}}) \cdot {(x_s^i W^k_{\operatorname{n}})}^T}{\sqrt{d_k}}\right)(x^i_s W^v_{\operatorname{n}}).
\end{aligned}
\end{equation}
Where $x_{m}^i$ represents the merged features of the $i$-th level sampled features. $C_u$ is the content embeddings updated by the positional embeddings. $W^O$ denotes the parameter matrices for multi-head concatenation. $W_n^q$, $W_n^k$ and $W_n^v$ denote the parameter matrices for query, key and value of n-th attention head. $F_{\operatorname{concat}}$ is the concatenating operation. $d_k=\frac{N_{hd}}{N_H}$, $N_{hd}$ is the hidden dimensions, and $N_H$ is the number of attention head.
\par
Following the merging of the sampled features at each scale based on spatial information, the merged features at each scale are first concatenated together as follows:
\begin{equation}
\begin{aligned}
\centering
 X_{m}=F_{\operatorname{concat}}(\{x_{m}^i\}_{i=0,1,2}) \in \mathbb{R}^{B \times N_q \times N_L \times N_{hd}},
\end{aligned}
\end{equation}
where $N_L$ is the number of multi-scale, $x_{m}^i$ represents the merged features of the $i$-th level sampled features, $X_m$ is the concatenated multi-scale features and merged by the scale-aware merging mechanism as follows: 
% \begin{equation}
% \begin{aligned}
% \centering
% X_{u}= F_{\text{\tiny MHA}}\left( C_u W^q,X_{m}W^k,X_{m}W^v\right) 
% % \in \mathbb{R}^{B \times N \times C}
% \end{aligned}
% \end{equation}
\begin{equation}
\begin{aligned}
X_{u} &=F_{\operatorname{concat}}\left(\operatorname{head}_1, \ldots, \operatorname{head}_{\mathrm{h}}\right) W^O, \\
 \operatorname{ where }\ \ {\operatorname{head}}_{\operatorname{n}} &=\operatorname{Softmax} \left(\frac{(C_u W^q_{\operatorname{n}}) \cdot {(X_m W^k_{\operatorname{n}})}^T}{\sqrt{d_k}}\right)(X_m W^v_{\operatorname{n}}).
\end{aligned}
\end{equation}
Where $X_u$ is the merged multi-scale features for updating the content embeddings.

\begin{table*}[htbp]
\caption{Instance statistics of two difficulties. We quantify all the instances in the HAKE-HOI \cite{r39} dataset according to two newly proposed metrics and divide them into ten intervals.}
\centering
 \resizebox{0.9\textwidth}{!}{
\begin{tabular}{cccccccccccc}
\toprule
                               Dataset  & IMI& \multicolumn{1}{c}{$\text{IMI}_0$}   & \multicolumn{1}{c}{$\text{IMI}_1$} & \multicolumn{1}{c}{$\text{IMI}_2$} & \multicolumn{1}{c}{$\text{IMI}_3$} & \multicolumn{1}{c}{$\text{IMI}_4$} & \multicolumn{1}{c}{$\text{IMI}_5$} & \multicolumn{1}{c}{$\text{IMI}_6$} & \multicolumn{1}{c}{$\text{IMI}_7$} & \multicolumn{1}{c}{$\text{IMI}_8$} & $\text{IMI}_9$    \\ \midrule
\multirow{2}{*}{HAKE-HOI}            & $\text{num}_{AR}$     & 104243 & 65499 & 44303 & 31241 & 21982 & 11888 & 4670  & 1818  & 598   & 168  \\
                                 & $\text{num}_{LR}$ & 424&	1243	&1784&	3043&	8668	&70191	&83314&	79427&	34017&	4299 \\ \midrule
\multirow{2}{*}{SDC\_Train} & $\text{num}_{AR}$     & 62526  & 30235 & 16346 & 12013 & 10269 & 11189 & 4223  & 1540  & 423   & 139  \\
                                 & $\text{num}_{LR}$ & 177    & 515   & 874   & 1656  & 5208  & 48798 & 38517 & 29544 & 20265 & 3349 \\ \midrule
\multirow{2}{*}{SDC\_Test}  & $\text{num}_{AR}$     & 24737  & 0     & 0     & 0     & 0     & 0     & 0     & 0     & 0     & 0    \\
                                 & $\text{num}_{LR}$ & 153    & 415   & 464   & 834   & 2704  & 20167 & 0     & 0     & 0     & 0    \\ \bottomrule
\end{tabular}}
\label{table 1}
\end{table*}

\subsubsection{Task-Aware Merging Mechanism}\label{sec.7}
Considering diverse HOI instances, the task-aware merging mechanism is proposed to fuse the merged multi-scale features and content embeddings and align the content embeddings with the merged feature in the task-aware perspective, as shown in Fig.\ref{fig:Figure3} (e). It leverages the merged multi-scale features and content embeddings to generate dynamic switch for selecting suitable channel in the merging process. Content embedding and multi-scale information after fusion are first stitched together, the formulation is as follows:
\begin{equation}
    \begin{aligned}
    \centering
    X= F_{stack}\left(C_u, X_u \right)  \in \mathbb{R}^{B \times N_q \times( 2 \times N_{hd})}.
    \end{aligned}
\end{equation}
Where  $C_u$ is the content embeddings updated by the positional embeddings, $X_u$ is the merged multi-scale features. Thereafter, we use cross-attention mechanism to update these as follows:
\begin{equation}
    \begin{aligned}
    X_{switch} &=F_{\operatorname{concat}}\left(\operatorname{head}_1, \ldots, \operatorname{head}_{\mathrm{h}}\right) W^O, \\
     \operatorname{ where }\ \ {\operatorname{head}}_{\operatorname{n}} &=\operatorname{Softmax} \left(\frac{(C_u W^q_{\operatorname{n}}) \cdot {(X W^k_{\operatorname{n}})}^T}{\sqrt{d_k}}\right)(X W^v_{\operatorname{n}}).
    \end{aligned}
\end{equation}
Then, the generated information is utilized to gain the dynamic switch for merging, the formulation is as follows:
\begin{equation}
    \begin{aligned}
    \centering
    Switch^{\gamma}= F_{normalize}(F_{mlp}(X_{switch}))^{\gamma} \in \mathbb{R}^{B \times N_q \times 2\times 2},
    \end{aligned}
\end{equation}
where $Switch^{\gamma}$ is the dynamic switch for ${\gamma}$-th dimension of the merged features. $F_{hsigmoid}(\cdot)$ and $F_{mlp}(\cdot)$ denote the hard sigmoid and feed forward network which consists of two linear layers and one Relu activation layer, respectively. Inspired by \cite{r69}, the merging mechanism is designed as follows:
\begin{equation}
    \begin{aligned}
    \centering
    U^{\gamma}= F_{Max}\{Switch^{\gamma}_{i,0} \odot X_u^{\gamma} + Switch^{\gamma}_{i,1}\}_{i=0,1} + C_u^{\gamma},
    % \in \mathbb{R}^{B \times N \times 1}
    \end{aligned}
\end{equation}
where $U^{\gamma}$ is $\gamma$-th features of content embeddings updated by the merged multi-scale features. $F_{Max}$ is the max operation.

\begin{figure}[htbp]
    \centering
    \includegraphics[width =\linewidth]{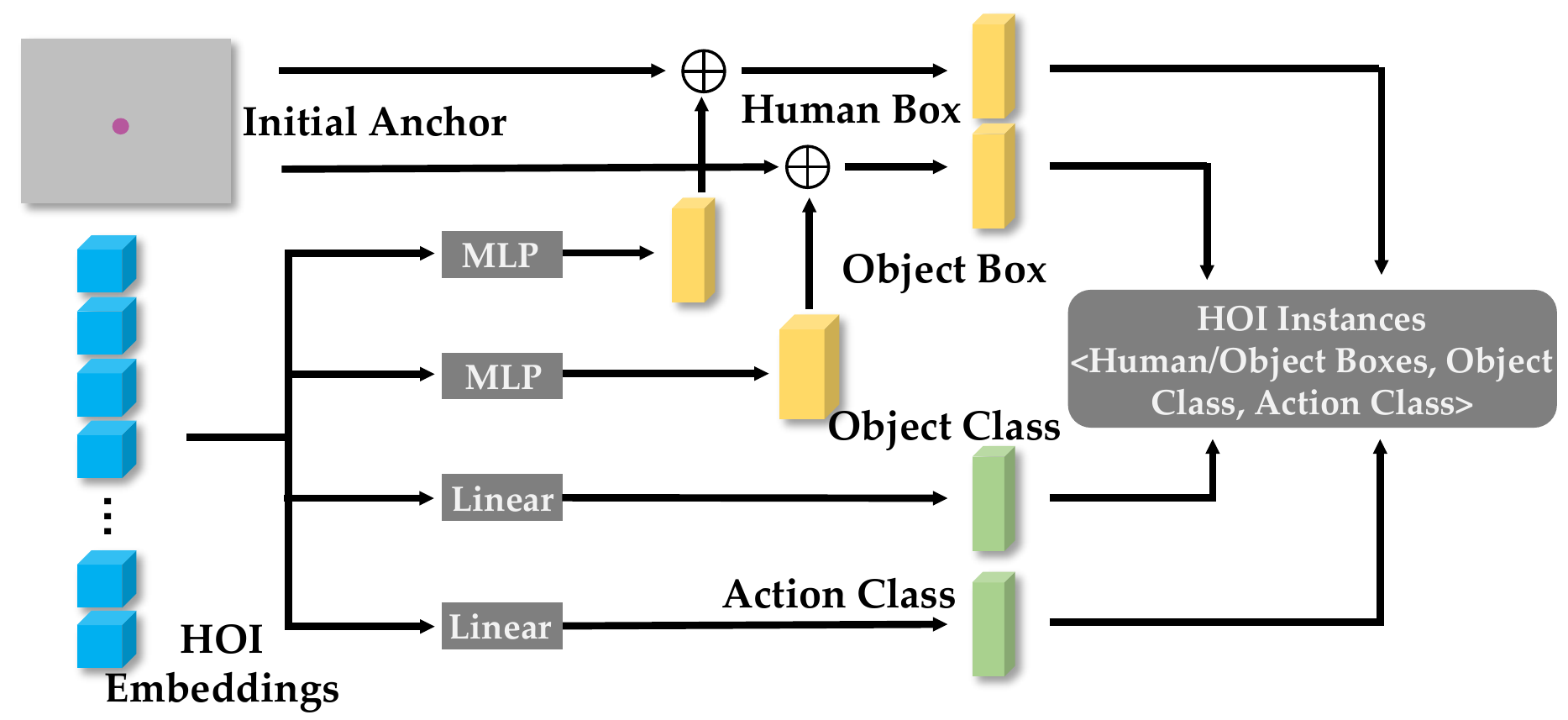}
    \caption{The prediction process of the HOI detection head. See sec \ref{sec.9} for more details.}
    \label{fig:Figure4}
\end{figure}

\begin{figure*}[htbp]
    \centering
    \includegraphics[scale=0.6]{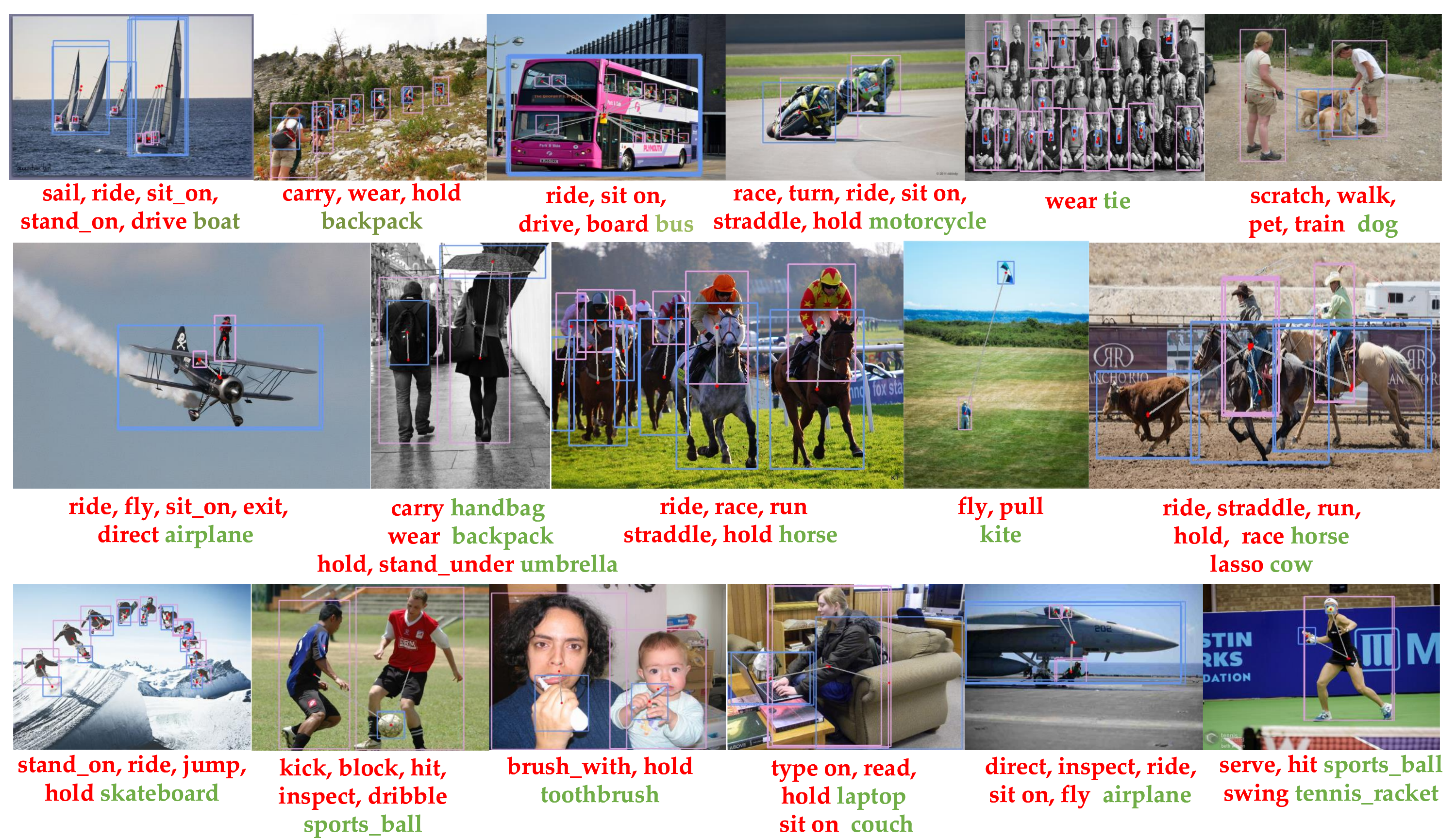}
    \caption{Visualization of HOI detection. Humans and objects are represented by pink and blue bounding boxes respectively, and interactions are marked by grey lines linking the box centers. Kindly refer to Sec. \ref{visualized_results} for more details.}
    \label{fig:Figure5}
\end{figure*}
 
\begin{figure}[htbp]
    \centering
    \includegraphics[width = \linewidth]{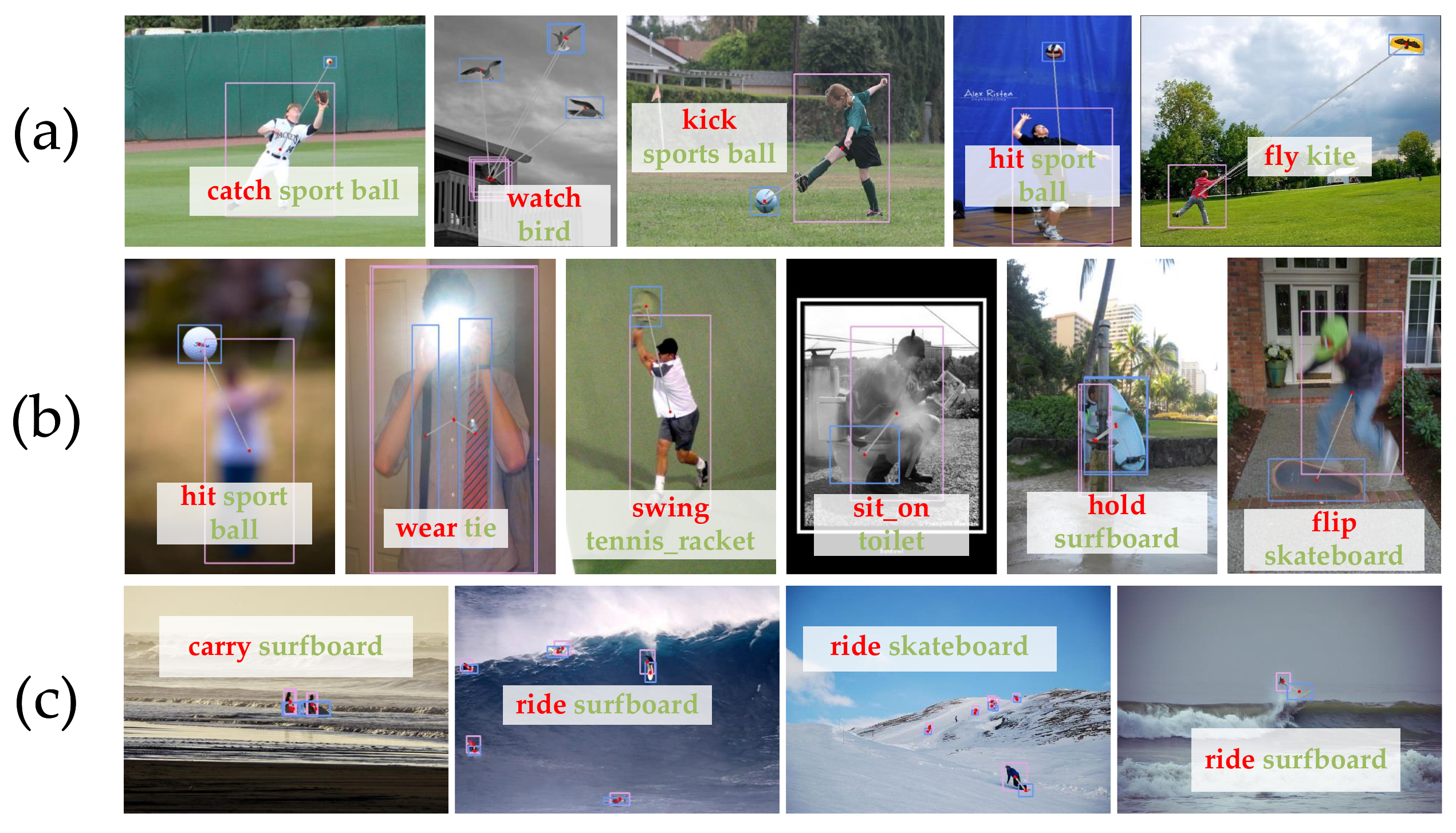}
    \caption{(a) illustrates the excellent long-range visual modelling capabilities. (b) demonstrates remarkable robustness. (c) shows the superior capabilities for identifying small HOI instances. Kindly refer to Sec. \ref{visualized_results} for more details.}
    \label{fig:Figure6}
\end{figure}
\subsubsection{Decoding with Fine-Grained Anchor}\label{sec.8}
As shown in Fig.\ref{fig:Figure3} (e), the updated content embeddings are used to generate fine-grained anchors and attention weights. According to the linear layer, reshape operation and softmax function, the formulation is as follows:
\begin{equation}
\begin{aligned}
\mathcal{A} = F_{lin\&res}(U) \in \mathbb{R}^{B \times N_q \times N_H\times N_L\times N_{\mathcal{A}} \times 2},
\end{aligned}
\end{equation}
\begin{equation}
\begin{aligned}
\mathcal{W} = F_{lin\&res\&soft}(U) \in \mathbb{R}^{B \times N_q \times N_H\times N_L\times N_{\mathcal{A}} },
\end{aligned}
\end{equation}
As shown in Fig.\ref{fig:Figure3} (a), the fine-grained anchors and attention weights are utilized to aid semantic features from the encoded features of the input scenarios to the content embeddings, the formulation is as follows:
\begin{equation}
\begin{aligned}
\mathcal{P}_{q} =\sum_{n=1}^{N_{H}} \boldsymbol{W}_n\left[\sum_{l=1}^{N_{L}} \sum_{k=1}^{N_{\mathcal{A}}} \mathcal{W}_{nqk}^l \cdot \boldsymbol{W}_n^{\prime} \boldsymbol{x^l}\left(\mathcal{A}_{nqk}^l\right)\right],
\end{aligned}
\end{equation}
where $\mathcal{P}_{q}$ is the extracted semantic information used for translating $q$-th content to HOI embeddings. $\mathcal{A}_{nqk}^l$ and $\mathcal{W}_{nqk}^l$  represent 
the $k$-th fine-grained anchors and corresponding attention weights of the $n$-th attention head for the $q$-th query embedding. Both $W_n$ and $W_n^{\prime}$ are parameter matrices of the $n$-th attention head.
$N_{\mathcal{A}}$ is the number of fine-grained anchors of each scale in one attention head.

\subsection{HOI Detection Head}\label{sec.9}
FGAHOI leverages a simple HOI detection head to predict all elements of HOI instances. As shown in Fig.\ref{fig:Figure4}, the detection head utilizes the HOI embeddings and the initial anchor to localize the human and object boxes. In this process, each initial anchor acts as the base point for the bounding boxes of the corresponding pair of a human and an object, the formulation is as follows:
\begin{equation}
b_h=F_{mlp}(H)[\cdots,:2]+initial\ anchor\ \ \in \mathbb{R}^{N_q \times 4}, 
\end{equation}
\begin{equation}
b_o=F_{mlp}(H)[\cdots,:2]+initial\ anchor\ \ \in \mathbb{R}^{N_q \times 4}, 
\end{equation}
\begin{equation}
c_o=F_{linear}(H)\ \ \in \mathbb{R}^{N_q \times num_o}, 
\end{equation}
\begin{equation}
c_v=F_{linear}(H)\ \ \in \mathbb{R}^{N_q \times num_v}, 
\end{equation}
where $F_{mlp}$ denotes the feed forward network consists of three linear layers and three relu activation layers. $F_{linear}$ stands for the linear layer. $num_o$ and $num_v$ are the number of object and action classes, respectively. $H$ denotes the HOI embeddings.

\begin{figure*}[htbp]
    \centering
    \includegraphics[width = \textwidth]{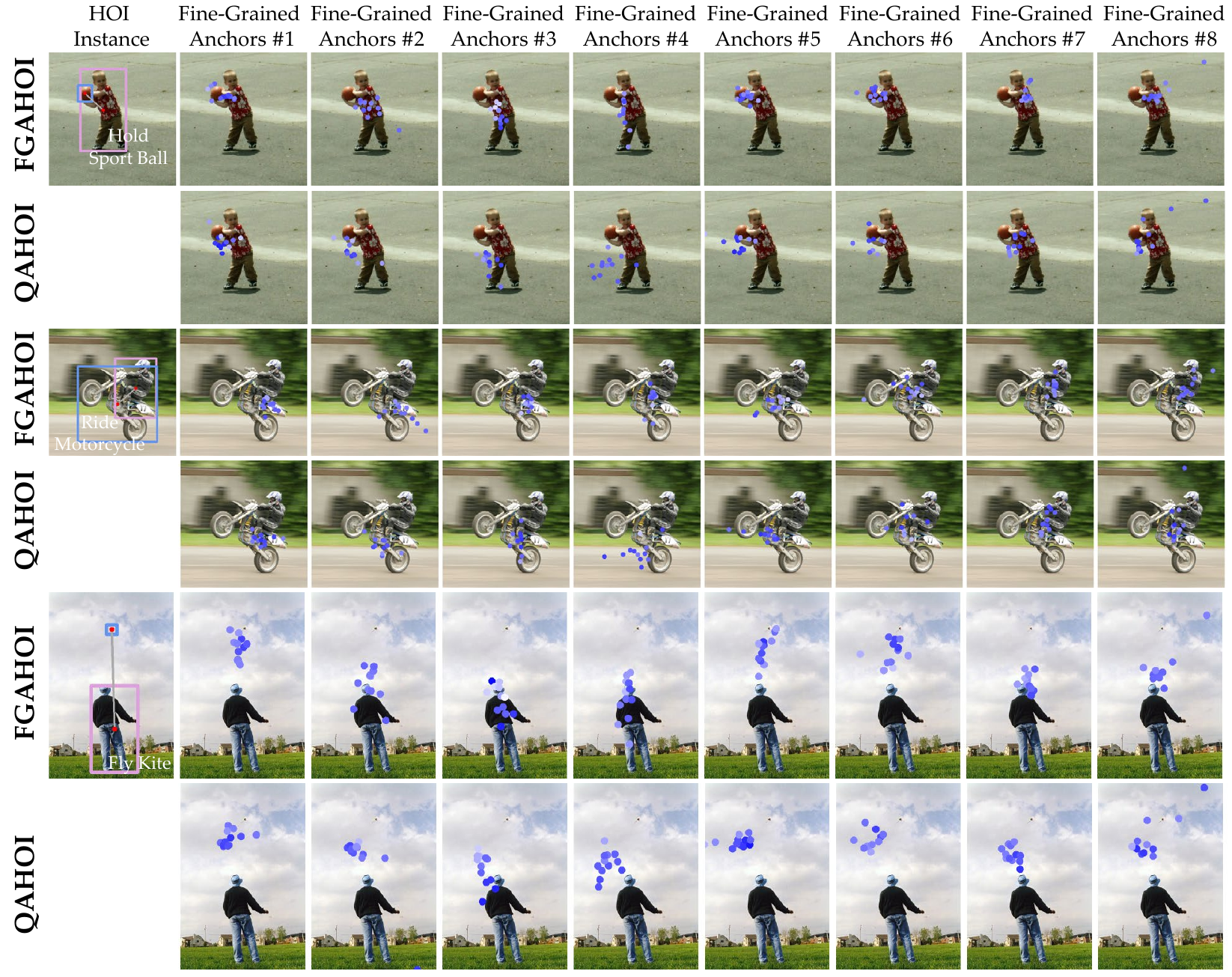}
    \caption{Comparison of fine-grained anchors between FGAHOI and QAHOI. We visualize the fine-grained anchors corresponding to all attention heads and the corresponding attention weights, where the shades of colors correspond to the magnitude of the weights. Obviously, FGAHOI is more accurate in focusing on humans, objects and interaction areas. Kindly refer to Sec. \ref{FGA_lookat} for more details.}
    \label{fig:Figure7}
\end{figure*}

\subsection{Training and Inference}\label{sec.10}

\subsubsection{Stage-wise Training}
Inspired by the stage-wise training approach \cite{r73,r74} which decouples feature learning and classifier learning into two independent stages for LTR \cite{r75}, we propose a novel stage-wise training strategy for FGAHOI. We start by training the base network (FGAHOI without any merging mechanism) in an end-to-end manner. We then add the merging mechanism in turn to the trained base network for another short period of training. In this phrase, the parameters of the trained base network are leveraged as pretrained parameters and no parameters are fixed during the training process. 

\subsubsection{Loss Calculation} Inspired by the set-based training process of HOI-Trans \cite{r21}, QPIC \cite{r33}, CDN \cite{r20} and QAHOI \cite{r34}, we first use the bipartite matching with the Hungarian algorithm to match each ground truth with its best-matching prediction. For subsequent back-propagation, a loss is then established between the matched predictions and the matching ground truths. The folumation is as follows:
\begin{equation}
L=\lambda_o L_c^o+\lambda_v L_c^v + \sum_{k \in(h, o)}\left(\lambda_b L_b^k+\lambda_{G I o U} L_{G I o U}^k\right),
\end{equation}
where $L_c^o$ and $L_c^v$ represent the object class and action class loss, respectively. We utilize the modified focal loss function \cite{r66} and sigmoid focal loss function \cite{r65} for $L_c^v$ and $L_c^o$, respectively. $L_b$ is the box regression loss and consists of the $L1$ Loss. $L_{GIOU}$ denotes the intersection-over-union loss, the same as the function in QPIC \cite{r33}.  $\lambda_o$, $\lambda_v$, $\lambda_b$ and $\lambda_{G I o U}$ are the hyper parameters for adjusting the weights of each loss.
\subsubsection{Inference} The inference process is to composite the output of the HOI detection head to form HOI triplets. Formally, the $i$-th output prediction is generated as $<b_i^h,b_i^o,argmax_kc_i^{hoi}(k)>$. The HOI triplet score $c_i^{hoi}$ combined by the scores of action $c_i^v$ and object $c_i^o$ classification, formularized as $c_i^{hoi}=c_i^v \cdot c_i^o$.

% \begin{table}[htbp]
% \caption{Instance statistics of two difficulties. We quantify all the instances in the HAKE-HOI \cite{r39} dataset according to two newly proposed metrics and divide them into ten intervals.}
% \centering
% \label{statics}
% \resizebox{0.9\linewidth}{!}{
% \begin{tabular}{cccccc}
% \toprule
% IMI                    & \multicolumn{1}{c}{$\text{IMI}_0$}   & \multicolumn{1}{c}{$\text{IMI}_1$} & \multicolumn{1}{c}{$\text{IMI}_2$} & \multicolumn{1}{c}{$\text{IMI}_3$} & \multicolumn{1}{c}{$\text{IMI}_4$} \\ 
% \toprule
% \toprule
% \multicolumn{1}{l}{$\text{num}_{AR}$} & 104243                      & 65499                       & 44303                       & 31241                       & 21982                       \\
% $\text{num}_{LR}$                     & 186                         & 1062                        & 1673                        & 2808                        & 5351                        \\\toprule
% IMI                    & \multicolumn{1}{c}{$\text{IMI}_5$} & \multicolumn{1}{c}{$\text{IMI}_6$} & \multicolumn{1}{c}{$\text{IMI}_7$} & \multicolumn{1}{c}{$\text{IMI}_8$} & $\text{IMI}_9$                       \\
% \toprule
% \toprule
% \multicolumn{1}{l}{$\text{num}_{AR}$} & 11888                       & 4670                        & 1818                        & 598                         & 168                         \\
% $\text{num}_{LR}$                     & 61956                       & 86528                       & 83140                       & 32941                       & 4765                        \\ \toprule
% \end{tabular}}
% \end{table}
\begin{table*}[htbp]
\centering
\caption{Performance comparison with the state-of-the-art methods on the HICO-DET dataset. 'V', 'S', 'P' and 'L' represent the visual feature, spatial feature, human pose feature and language feature respectively. Fine-tuned Detection means the parameter of the model is pre-trained on the MS-COCO dataset. Backbone with '*' and '+' means that they are pre-trained on ImageNet-22K with 384×384 input resolution. QAHOI(R) represents that the results are reproduced on the same machine with our model. Kindly refer to Sec. \ref{sec.sota-hoico-det}  for more details.} 
% \normalsize
\label{table 2}
\resizebox{\textwidth}{!}{
\begin{tabular}{c l l c c c c c c c c}
\toprule
\multicolumn{1}{c}{\multirow{2}{*}{ \textbf{Architecture}}} &
\multirow{2}{*}{\textbf{Method}}&\multirow{2}{*}{ \textbf{Backbone}} & 
\multirow{2}{*}{ \textbf{Fine-tuned}} & 
\multirow{2}{*}{ \textbf{Feature}}&
\multicolumn{3}{c}{ \textbf{Default} ($\uparrow$)} &
\multicolumn{3}{c}{ \textbf{Known Object} ($\uparrow$)}\\[0.5ex]

\multicolumn{1}{c}{} & & & & &
\multicolumn{1}{c}{ \textbf{Full}} &  \textbf{Rare} &  \textbf{Non-Rare} &  \textbf{Full} &  \textbf{Rare} &
\multicolumn{1}{c}{ \textbf{Non-Rare}}\\

\hline
% \multirow{3}{*}{\multicolumn{11}{c}{Two-Stage Methods}}\\

\multicolumn{11}{c}{\multirow{2}{*}{\large\textbf{Two-Stage Methods}}} \\ 
\multicolumn{11}{c}{\multirow{4}{*}{}} \\ 
% \hline

\multirow{5}{*}{Multi-stream} & No-Frill\cite{r38} & ResNet-152 & \ding{55} & A+S+P & 17.18 & 12.17 & 18.08 & - & - & -\\[0.5ex]
\multirow{1}{*}{} & PMFNet \cite{r43} & ResNet-50-FPN &\ding{55} & A+S &	17.46 & 15.65 & 18.00 & 20.34 & 17.47 & 21.20\\[0.5ex]
\multirow{1}{*}{} & ACP \cite{r44} & ResNet-101 & \ding{52} & A+S+L & 21.96 & 16.43 & 23.62 & - & - & -\\[0.5ex]
\multirow{1}{*}{} & PD-Net \cite{r8} & ResNet-152 &\ding{55} & A+S+P+L & 22.37 & 17.61 & 23.79 & 26.86 & 21.70 & 28.44\\[0.5ex]
\multirow{1}{*}{} & VCL \cite{r4} & ResNet-50	 &\ding{52} & A+S & 23.63 & 17.21 & 25.55 & 25.98 & 19.12 & 28.03\\[1ex]

% \multicolumn{11}{c}{} \\

\multirow{4}{*}{Graph-Based} & RPNN \cite{r5} & ResNet-50 &\ding{55} & A+P & 17.35 & 12.78 & 18.71 & - & - & -\\[0.5ex]
\multirow{1}{*}{} & VSGNet \cite{r14} & ResNet-152 &\ding{55} & A+S & 19.80 & 16.05 & 20.91 & - & - & -\\[0.5ex]
\multirow{1}{*}{} & DRG \cite{r10} & ResNet-50-FPN &\ding{52} & A+S+L & 24.53 & 19.47 & 26.04 & 27.98 & 23.14 & 29.43\\[0.5ex]
\multirow{1}{*}{} & SCG \cite{r24} & ResNet-50-FPN &\ding{52} & A+S & 31.33 & 24.72 & 33.31 & 34.37 & 27.18 & 36.50\\[0.5ex]

% \multicolumn{11}{c}{} \\

% \multirow{3}{*}{Transformer-Based} &
% \multirow{3}{*}{UPT \cite{r58}} & ResNet-50 &\ding{52} & A & 31.66 & 25.94 & 33.36 & 35.05 & 29.27 & 36.77\\
% \multirow{1}{*}{} & & ResNet-101 &\ding{52} & A & 32.31 & 28.55 & 33.44 & 35.65 & \textbf{31.60} & 36.86\\
% \multirow{1}{*}{} & & ResNet-101-DC5 &\ding{52} & A & 32.62 & \textbf{28.62} & 33.81 & 36.08 & 31.41 & 37.47\\[1.5ex]

\hline

\multicolumn{11}{c}{\multirow{2}{*}{\large\textbf{One-Stage Methods}}} \\ 
\multicolumn{11}{c}{\multirow{4}{*}{}} \\ 
\multirow{3}{*}{Interaction points} & IP-Net \cite{r19} & ResNet-50-FPN &\ding{55} & A & 19.56 & 12.79 & 21.58 & 22.05 & 15.77 & 23.92\\[0.5ex]
\multirow{1}{*}{} & PPDM \cite{r45} & Hourglass-104 &\ding{52} & A & 21.73 & 13.78 & 24.10 & 24.58 & 16.65 & 26.84\\[0.5ex]
\multirow{1}{*}{} & GGNet \cite{r9} & Hourglass-104 &\ding{52} & A & 23.47 & 16.48 & 25.60 & 27.36 & 20.23 & 29.48\\[1.5ex]
% \multicolumn{11}{c}{} \\

\multirow{15}{*}{Transformer-Based} & HOITrans\cite{r21} & ResNet-101 &\ding{52} & A & 26.60 & 19.15 & 28.54 & 29.1 & 20.98 & 31.57\\[0.5ex]

\multirow{2}{*}{} &
\multirow{2}{*}{HOTR\cite{r7}}& ResNet-50 &\ding{55} & A & 23.46 & 16.21 & 25.65 & - & - & -\\
\multirow{2}{*}{} & & ResNet-50 &\ding{52} & A & 25.10 & 17.34 & 27.42 & - & - & -\\[0.5ex]

\multirow{2}{*}{} &
\multirow{2}{*}{AS-Net\cite{r2}}& 
ResNet-50 & \ding{55} & A & 24.40 & 22.39 & 25.01 & 27.41 & 25.44 & 28.00\\
\multirow{2}{*}{} & & ResNet-50 & \ding{52} & A & 28.87 & 24.25 & 30.25 & 31.74 & 27.07 & 33.14\\[0.5ex]

\multirow{2}{*}{} &
\multirow{2}{*}{QPIC\cite{r33}}& ResNet-50 & \ding{52} & A & 29.07 & 21.85 & 31.23 & 31.68 & 24.14 & 33.93\\
\multirow{1}{*}{} & & ResNet-50 & \ding{55} & A  & 24.21 & 17.51 & 26.21 & - & - & -\\[0.5ex]

% \multirow{1}{*}{} & QPIC \cite{r33}&ResNet-50 & \ding{52} & A & 29.07 & 21.85 & 31.23 & 31.68 & 24.14 & 33.93\\[0.5ex]
% \multirow{1}{*}{} & CDN-S \cite{r20}&ResNet-50 & \ding{52} & A & 31.44 & 27.39 & 32.64 & 34.09 & 29.63 & 35.42\\
% \multirow{1}{*}{} & CDN-B \cite{r20}&ResNet-50 & \ding{52} & A & 31.78 & 27.55 & 33.05 & 34.53 & 29.73 & 35.96\\
% \multirow{1}{*}{} & CDN-L \cite{r20}&ResNet-101 & \ding{52} & A & 32.07 & 27.19 & 33.53 & 34.79 & 29.48 & 36.38\\[0.5ex]

\multirow{2}{*}{} &
\multirow{2}{*}{QAHOI\cite{r34}}& Swin-Tiny & \ding{55} & A & 28.47 & 22.44 & 30.27 & 30.99 & 24.83 & 32.84\\
\multirow{1}{*}{} & & $\text{Swin-Large}^*_+$ & \ding{55} & A  & 35.78 & 29.80 & 37.56 & 37.59 & 31.66 & 39.36\\[0.5ex]

\multirow{2}{*}{} &
\multirow{2}{*}{QAHOI (R)}& Swin-Tiny & \ding{55} & A & 27.67 & 20.22 & 29.69 & 30.06& 22.95 & 32.18 \\
\multirow{1}{*}{} & & $\text{Swin-Large}^*_+$ & \ding{55} & A  & 35.43 & 29.22 & 37.29 & 37.23 & 31.01 & 39.09\\[0.5ex]

\multirow{2}{*}{} &
\multirow{2}{*}{FGAHOI}& Swin-Tiny & \ding{55} & A & \textbf{29.94}   & \textbf{22.24}   & \textbf{32.24}    & \textbf{32.48}    & \textbf{24.16}    & \textbf{34.97} \\
\multirow{1}{*}{} & & $\text{Swin-Large}^*_+$ & \ding{55} & A & \textbf{37.18} & \textbf{30.71} & \textbf{39.11} & \textbf{38.93} & \textbf{31.93} & \textbf{41.02}\\[0.5ex]

\toprule
\end{tabular}}
\end{table*}

\section{Proposed Dataset}
There are two main difficulties existing with human-object pairs. $i)$ Uneven size distribution of human and objects in human-object pairs. $ii)$ Excessive distance between person and object in human-object pairs. To the best of our knowledge, there are no relevant metrics to measure these two difficulties. In this paper, we propose two metrics $AR$ and $LR$ for measuring these two difficulties. Then two novel challenges corresponding to these two difficulties are proposed. In addition, we propose a novel \textbf{S}et for these \textbf{D}ouble \textbf{C}hallenges (HOI-SDC). The data is selected from HAKE-HOI \cite{r39} which is re-split from HAKE \cite{r67} and provides 110K+ images. HAKE-HOI has 117 action classes, 80 object classes and 520 HOI categories.

\subsection{HOI-UDA} \label{sec.HOI-UDA}

We propose a novel measurement for the challenge of \textbf{Uneven Distributed Area in Human-Object Pairs}, the formulation is as follow:
\begin{equation}
AR=\frac{{Area}_h \cdot {Area}_o}{{Area}_{hoi}^2},
\end{equation}
where ${Area}_h$, ${Area}_o$ and ${Area}_{hoi}$ denote the area of human, object and HOI instances, respectively (as shown in Fig.\ref{fig:Figure8} (a)). We quantify all the instances in the HAKE-HOI into ten intervals and count the number of instances of each interval in the second and fifth row of Table.\ref{table 1}. To better evaluate the ability of the model to detect HOI for human-object pairs with uneven distributed areas, we specially select 24737 HOI instances of $\text{IMI}_0^\text{UDA}$ in testing set.
\subsection{HOI-LDVM} \label{sec.HOI-LDVM}

\begin{figure}[htbp]
    \centering
    \includegraphics[width = \linewidth]{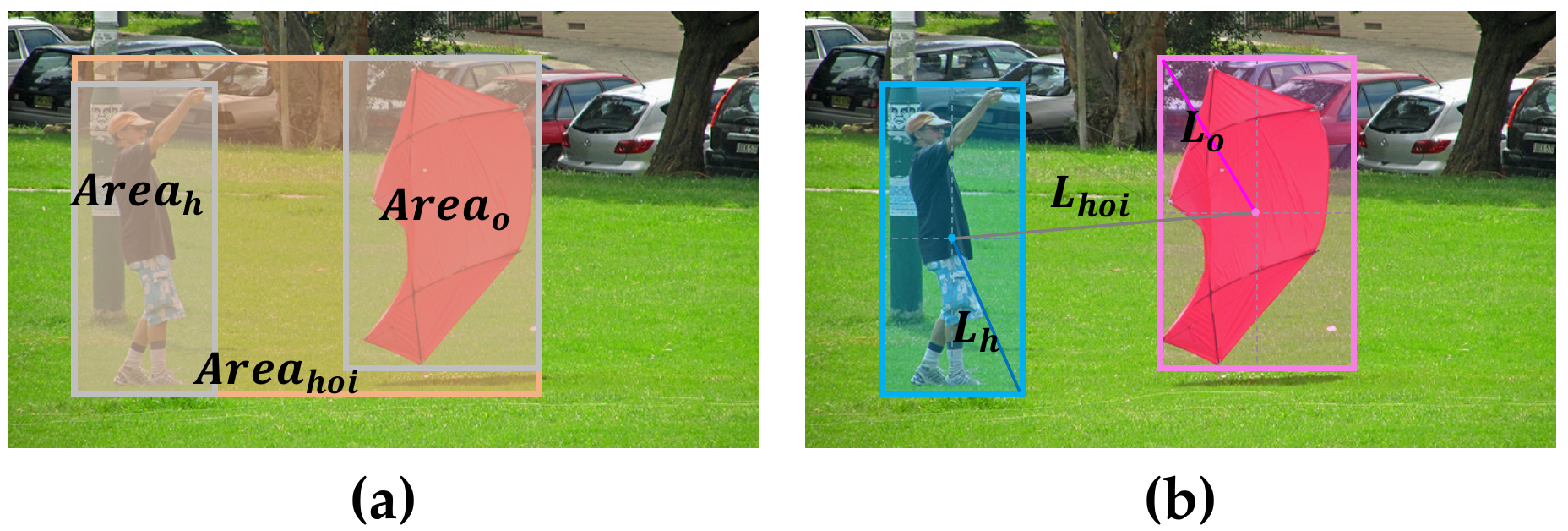}
    % \DeclareGraphicsExtensions.
    \caption{Proposed metrics for the difficulties existing with HOI instances. (a) Metric for uneven size distribution of humans and objects. (b) Metric for excessive distance between person and object. Kindly refer to Sec. \ref{sec.HOI-UDA} and \ref{sec.HOI-LDVM} for more details.}
    \label{fig:Figure8}
\end{figure}

A novel measurement for the challenge of \textbf{Long Distance Visual Modeling of Human-Object Pairs} is proposed in Eq.\ref{LDVM}. 
\begin{equation}
 LR=\frac{{L}_h + {L}_o}{{L}_{hoi}},
\label{LDVM}
\end{equation}
where ${L}_h$, ${L}_o$ and ${L}_{hoi}$ denote the size we define of human, object and HOI instances, respectively (as shown in Fig.\ref{fig:Figure8} (b)). The instances are quantified in the third and sixth row of Table.\ref{table 1}.  To better evaluate the ability of the model to detect HOI for human-object pairs with with long distance, we specially select 24737 HOI instances of $\text{IMI}_0^\text{LDVM} \sim \text{IMI}_6^\text{LDVM}$ in testing set.
\subsection{HOI-SDC}
In order to avoid the training process of the model being influenced by a portion of HOI classes with a very small number of instances, we remove some of the HOI classes containing a very small number of instances and HOI classes with no interaction from the training \textbf{S}et for the \textbf{D}ouble \textbf{C}hallenge. Finally, there are total 321 HOI classes, 74 object classes and 93 action classes. The training and testing set contain 37,155 and 9,666 images, respectively. The detailed distribution of HOI instances is shown in Table.\ref{table 1}.
 
% Table 1
\begin{table}[htbp]
\setlength{\tabcolsep}{12pt}
\centering
\caption{Performance comparison with the state-of-the-art methods on the HOI-SDC dataset.  Kindly refer to Sec. \ref{HOI-SDC} for more details.}
\label{table 4}
% \resizebox{0.8\linewidth}{!}{
\begin{tabular}{cccccccc}
\toprule
% \multirow{2}{*}{Method} & \multicolumn{7}{c}{$\text{mAP}_{\text{role}}$ ($\uparrow$)}      \\
%                         & $\text{IMI}_{\text{all}}$ & $\text{IMI}_0$ & $\text{IMI}_1$ & $\text{IMI}_2$ & $\text{IMI}_3$ & $\text{IMI}_4$& $\text{IMI}_5$ \\ \midrule
% QAHOI                   & 19.55    & 2.88 & 6.57 &  10.88 & 10.21  &  13.37 & 20.99   \\[1ex]
% Baseline          & 21.18    & 4.59 & 8.93& 9.97  &  13.07 &  15.49 & 22.87  \\[1ex]
% +HSAM        & 22.17    & 4.45 & 7.28 & 8.96  & 12.74  & 14.76  &  23.65 \\[1ex]
% +TAM      & 21.84   & 3.77 &5.57 &  9.92 & 12.68  & 15.25  & 23.53  \\[1ex]
% FGAHOI                   & -    & - & - &   &   &   &   \\\bottomrule
\multicolumn{2}{c}{Dataset} &\multicolumn{2}{c}{Backbone}  & \multicolumn{2}{c}{Method}   & \multicolumn{2}{c}{$\text{mAP}_{\text{role}}$ ($\uparrow$)} \\ \midrule
\multicolumn{2}{c}{\multirow{7}{*}{HOI-SDC}} & \multicolumn{2}{c}{Swin-Tiny}& \multicolumn{2}{c}{QAHOI}    & \multicolumn{2}{c}{19.55}   \\[1ex]
\multicolumn{2}{c}{}                          & \multicolumn{2}{c}{Swin-Tiny}& \multicolumn{2}{c}{Baseline} & \multicolumn{2}{c}{21.18}   \\[1ex]
\multicolumn{2}{c}{}                          & \multicolumn{2}{c}{Swin-Tiny}& \multicolumn{2}{c}{+HSAM}    & \multicolumn{2}{c}{21.91}   \\[1ex]
\multicolumn{2}{c}{}                         & \multicolumn{2}{c}{Swin-Tiny} & \multicolumn{2}{c}{+TAM}     & \multicolumn{2}{c}{21.84}   \\[1ex]
\multicolumn{2}{c}{}                          & \multicolumn{2}{c}{Swin-Tiny}& \multicolumn{2}{c}{FGAHOI}   & \multicolumn{2}{c}{22.25}   \\ \bottomrule

\end{tabular}%}
\end{table}

\begin{table}[htbp]
\centering
\caption{Performance comparison with the state-of-the-art methods on the V-COCO dataset. Kindly refer to Sec. \ref{V-COCO} for more details. }
\small
\label{table 3}
\resizebox{\linewidth}{!}{
\begin{tabular}{c l c c}
% \hline
\toprule
\multicolumn{1}{c}{} &\textbf{Method} &
$\mathrm{AP}_{\text {role }}^{S 1}$ ($\uparrow$) & $\mathrm{AP}_{\text {role }}^{S 2}$ ($\uparrow$)
\\[0.5ex]
\midrule
\multicolumn{1}{c}{\multirow{4}{*}{Two-stage Method}}&
\multirow{1}{*}{}VSG-Net &	51.8 &	57.0\\[0.5ex]
\multirow{1}{*}{} &PD-Net &	52.0 &	- 	\\[0.5ex]
\multirow{1}{*}{} &ACP &	53.2 &	- 	\\[0.5ex]

\midrule

\multicolumn{1}{c}{\multirow{9}{*}{One-stage Method}}&
\multirow{1}{*}{} HOITrans &52.9 &	-\\[0.5ex]
\multirow{1}{*}{} &AS-Net &	53.9 &	- 	\\[0.5ex]
\multirow{1}{*}{}& HOTR &55.2 &	64.4 	\\[0.5ex]
\multirow{1}{*}{} &DIRV &56.1 &	- 	\\[0.5ex]
\multirow{1}{*}{} &QAHOI(R-50) &58.2 &	58.7	\\[0.5ex]
\multirow{1}{*}{} & FGAHOI(R-50) &59.0 &	59.3 	\\[0.5ex]
\multirow{1}{*}{} & FGAHOI(Swin-T) &60.5 &	61.2 	\\[0.5ex]
\bottomrule
% \toprule
\end{tabular}}
\end{table}

\begin{table*}[htbp]
\centering
\caption{Comparison on ten intervals of the two proposed challenges. We divide the HICO-DET dataset into ten intervals based on each of the two challenges and compare the performance of QAHOI and FGAHOI on each interval.  Kindly refer to Sec. \ref{Sensitivity Analysis} for more details.}
\label{table 5}
\resizebox{0.9\textwidth}{!}{
\begin{tabular}{cllcccccccccc}
\toprule
\multirow{2}{*}{Challenge} & \multirow{2}{*}{Method} & \multirow{2}{*}{Backbone} & \multicolumn{10}{c}{$\text{mAP}_{\text{role}}$ ($\uparrow$)}                                                                \\[0.5ex]
                        &                         &                            & $\text{IMI}_0$ & $\text{IMI}_1$ & $\text{IMI}_2$ & $\text{IMI}_3$ & $\text{IMI}_4$& $\text{IMI}_5$     &  $\text{IMI}_6$      &  $\text{IMI}_7$      &  $\text{IMI}_8$      &  $\text{IMI}_9$      \\ \midrule
\multirow{4}{*}{UDA}     & \multirow{2}{*}{QAHOI}  & Swin-Tiny                 & 16.35 & 24.72 & 29.24 & 34.79 & 38.70  & 46.21 & 53.13 & 47.60  & 58.66 & 60.19 \\[0.5ex]
                        &                         & $\text{Swin-Large}^*_+$                & 20.53 & 33.58 & 41.11 & 45.41 & 45.44 & 56.43 & 56.25 & 63.53 & 71.12 & 75.08 \\[0.5ex]
                        & \multirow{2}{*}{FGAHOI} & Swin-Tiny                 & 19.74 & 29.85 & 32.20  & 39.46 & 40.54 & 48.55 & 51.32 & 46.50  & 66.44 & 78.17 \\[0.5ex]
                        &                         & $\text{Swin-Large}^*_+$              & 23.69 & 35.85 & 42.51 & 50.50  & 46.89 & 56.95 & 56.33 & 63.04 & 75.70  & 79.42 \\ \midrule
\multirow{4}{*}{LDVM}      & \multirow{2}{*}{QAHOI} & Swin-Tiny                 & 1.33 & 4.43 & 2.57 & 5.00   & 8.06 & 17.87 & 22.81 & 29.25 & 34.03 & 42.29 \\[0.5ex]
                        &                         & $\text{Swin-Large}^*_+$              & 0.82 & 4.08 & 2.56 & 7.53 & 11.42 & 22.87 & 30.94 & 41.38 & 45.31 & 60.15 \\[0.5ex]
& \multirow{2}{*}{FGAHOI}  & Swin-Tiny                 & 2.50  & 4.15 & 3.34 & 7.58 & 9.83 & 21.61 & 27.64 & 33.07 & 38.31 & 45.07 \\[0.5ex]
                        &                         & $\text{Swin-Large}^*_+$                & 1.44 & 4.32 & 4.57 & 7.81 & 11.82 & 24.92 & 32.50  & 43.66 & 47.26 & 60.55 \\
                        \bottomrule
\end{tabular}}
\end{table*}

\begin{table*}[htbp]
\centering
\caption{We carefully ablate each of the constituent component of FGAHOI. The middle results denote the role mAP. The results in the top right corner represent the performance improvement compared to QAHOI. The results in the bottom right corner represent the performance improvement compared to the baseline. Kindly refer to Sec. \ref{Ablating FGAHOI} for more details.}
\renewcommand\arraystretch{1.15}

\label{table 6}

\begin{tabular}{ccccccccc}
\toprule
\multirow{2}{*}{\textbf{Method}}                      & \multicolumn{2}{c}{\textbf{Merging Mechanism}}                                                     & \multicolumn{3}{c}{\textbf{Default}}  & \multicolumn{3}{c}{\textbf{Known Object}} \\ & \textbf{Hierarchical Spatial-Aware}            & \textbf{Task-Aware}      & \textbf{Full $\uparrow$}    & \textbf{Rare $\uparrow$}    & \textbf{Non-Rare $\uparrow$} & \textbf{Full $\uparrow$}      & \textbf{Rare $\uparrow$}     & \textbf{Non-Rare $\uparrow$}  \\ \midrule
 %----------------------------------------
\multicolumn{1}{c|}{QAHOI}                   & -                            & \multicolumn{1}{c|}{-}                  & 27.67   & 20.22   & 29.69    & 30.06         & 22.95  & 32.18   \\ \midrule
 %----------------------------------------
\multicolumn{1}{c|}{\multirow{8}{*}{FGAHOI}}  & \multirow{1}{*}{\ding{55}} & \multicolumn{1}{c|}{\multirow{1}{*}{\ding{55}}} & $\text{28.45}_\text{(\ \ \ \ \ -\ \ \ \ \ )}^\text{( +0.78 )}$   & $\text{21.07}_\text{(\ \ \ \ \ -\ \ \ \ \ )}^\text{( +0.85 )}$   & $\text{30.66}_\text{(\ \ \ \ \ -\ \ \ \ \ )}^\text{( +0.97 )}$    & $\text{31.08}_\text{(\ \ \ \ \ -\ \ \ \ \ )}^\text{( +1.02 )}$       & $\text{24.02}_\text{(\ \ \ \ \ -\ \ \ \ \ )}^\text{( +1.01 )}$   & $\text{33.19}_\text{(\ \ \ \ \ -\ \ \ \ \ )}^\text{( +1.07 )}$    \\[3ex]
% \multicolumn{1}{c|}{}                        &                                        & \multicolumn{1}{c|}{}                   & {  ( +0.78 )}&	 { ( +0.85 )}&	 {  ( +0.97 )}& {  ( +1.02 )}& {  ( +1.01 )}	&{  ( +1.07 )}  \\[0.5ex] 
% \multicolumn{1}{c|}{}                        &                                  & 
% \multicolumn{1}{c|}{} & {  ( - )}&	 {  ( - )}&	 { ( - )}& {  ( - )}& {  ( - )}	&{ ( - )}\\[0.5ex] 
 %----------------------------------------
\multicolumn{1}{c|}{}                        & \multirow{1}{*}{\ding{52}} & \multicolumn{1}{c|}{\multirow{1}{*}{\ding{55}}} & $\text{29.60}^\text{( +1.93 )}_\text{( +1.15 )}$       & \textbf{$\text{22.39}^\text{( +2.17 )}_\text{( +1.32 )}$}         &  $\text{31.76}^\text{( +2.07 )}_\text{( +1.10 )}$       &  $\text{ 32.07}^\text{( +2.01 )}_\text{( +0.99 )}$        &  \textbf{$\text{24.48}^\text{( +1.53 )}_\text{( +0.46 )}$}     & $\text{34.34}^\text{( +2.16 )}_\text{( +1.15 )}$     \\[3ex]
% \multicolumn{1}{c|}{}                        &                                       & \multicolumn{1}{c|}{}                   & { ( +1.93 )}&	 { ( +2.17 )}&	 {  ( +2.07 )}& { ( +2.01 )}& {  ( +1.53 )}	&{  ( +2.16 )}       \\[0.5ex]
% \multicolumn{1}{c|}{}                        &                                   & \multicolumn{1}{c|}{}           &       { ( +1.15 )}&	 { ( +1.32 )}&	 {  ( +1.10 )}& { ( +0.99 )}& {  ( +0.46 )}	&{ ( +1.15 )}\\[0.5ex] 
 %---------------------------------------
\multicolumn{1}{c|}{}                        & \multirow{1}{*}{\ding{55}}  & \multicolumn{1}{c|}{\multirow{1}{*}{\ding{52}}} & $\text{29.32}^\text{( +1.65 )}_\text{( +0.87 )}$   & $\text{22.34}^\text{( +2.12 )}_\text{( +1.27 )}$   & $\text{31.41}^\text{( +1.72 )}_\text{( +0.75)}$     & $\text{31.81}^\text{( +1.75 )}_\text{( +0.73)}$     & $\text{24.30}^\text{( +1.35 )}_\text{( +0.28)}$    & $\text{34.05}^\text{( +1.87 )}_\text{( +0.86)}$     \\[3ex]
% \multicolumn{1}{c|}{}                       &            & \multicolumn{1}{c|}{}           & { ( +1.65 )}&	 { ( +2.12 )}&	 {  ( +1.72 )}& { ( +1.75 )}& { ( +1.35 )}	&{ ( +1.87 )}    	\\[0.5ex]
% \multicolumn{1}{c|}{}                  &      & \multicolumn{1}{c|}{}            & {  ( +0.87 )}&{  ( +1.27 )}& { ( +0.75 )}& {  ( +0.73 )}& {  ( +0.28 )} &{ ( +0.86 )}      \\ [0.5ex]
 %---------------------Task-------------------
\multicolumn{1}{c|}{}                        & \multirow{1}{*}{\ding{52}} &  \multicolumn{1}{c|}{\multirow{1}{*}{\ding{52}}} & \textbf{$\text{29.94}^\text{( +2.27 )}_\text{( +1.49 )}$}   & $\text{22.24}^\text{( +2.02 )}_\text{( +1.17 )}$   & \textbf{$\text{32.24}^\text{( +2.55 )}_\text{( +1.58 )}$}    & \textbf{$\text{32.48}^\text{( +2.42 )}_\text{( +1.40 )}$}    & $\text{24.16}^\text{( +1.21 )}_\text{( +0.14 )}$    & \textbf{$\text{34.97}^\text{( +2.79 )}_\text{( +1.78 )}$}     \\[1ex]
% \multicolumn{1}{c|}{}                        &                                       & \multicolumn{1}{c|}{}                   & {  ( +2.27 )}&	 { ( +2.02 )}&	 {  ( +2.55 )}& { ( +2.42 )}& {(  +1.21 )}	&{ ( +2.79 )}           \\[0.5ex]
% \multicolumn{1}{c|}{}                        &                                    & \multicolumn{1}{c|}{}                   & { ( +1.49 )}&{  ( +1.17 )}& { ( +1.58 )}& {  ( +1.40 )}& {  ( +0.14 )}&{ ( +1.78 )}        \\ 
\bottomrule
\end{tabular}
\end{table*}
% \begin{table}[HTBP]
% \centering
% \caption{Performance comparison with the state-of-the-art methods on the HAKE-HOI dataset.}
% \small
% \begin{tabular}{cc}
% \hline
% \textbf{Method}               & \multicolumn{1}{c|}{$mAP$} \\ \hline
% \multicolumn{1}{l}{}iCAN &                          \\
%                      R+iCAN+D&                          \\
%                      &                          \\
%                      & 1                        \\ \hline
% \end{tabular}
% \end{table}

% Table 3

% \begin{table}[htbp]
% \centering
% \caption{Performance comparison with the state-of-the-art methods on the HOI-LDVM dataset.}
% \label{table 11}
% \resizebox{0.8\linewidth}{!}{
% \begin{tabular}{cccc}
% \toprule
% Dataset                  & Method & Backbone    & mAP \\ \toprule
% \multirow{4}{*}{HOI-LDVM} & QAHOI  & swin-tiny   & -   \\
%                          & FGAHOI  & swin-tiny   & -   \\
%                          & QAHOI  & swin\_large & -   \\
%                          & FGAHOI  & swin\_large & -   \\ \toprule
% \end{tabular}}
% \end{table}

% \usepackage{multirow}

\section{Experiments}
\subsection{Dataset}
Experiments are conducted on three HOI datasets: HICO-DET \cite{r54}, V-COCO \cite{r59} and HOI-SDC dataset \par
\textbf{HICO-DET} \cite{r54} has 80 object classes, 117 action classes and 600 HOI classes. HICO-DET offers 47,776 images with 151,276 HOI instances, including 38,118 images with 117,871 annotated instances of human-object pairs in the training set and 9658 images with 33,405 annotated instances of human-object pairs in the testing set. According to the number of these HOI classes, the 600 HOI classes in the dataset are grouped into three categories: Full (all HOI classes), Rare (138 classes with fewer than ten instances) and Non-Rare (462 classes with more than ten instances). Following HICO \cite{r56}, we consider two different evaluation settings (the results are shown in Table.\ref{table 2}: (1) Known object settings: For each HOI category (such as 'flying a kite'), the detection is only evaluated on the images that contain the target object category (such as 'kite'). The difficulty lies in the localization of HOI (e.g. human-kite pairs) and distinguishing the interaction (e.g. 'flying'). (2) Default setting: For each HOI category, the detection is evaluated on the whole test set, including images containing and without target object categories. This is a more challenging setting because we also need to distinguish background images (such as images without 'kite'). \par
\textbf{V-COCO} \cite{r59} contains 80 different object classes and 29 action categories and is developed from the MS-COCO dataset, which includes 4,946 images for the test subset, 2,533 images for the train subset and 2,867 images for the validation subset. The objects are divided into two types: “object” and “instrument”. \par 
\subsection{Metric} \label{Metric}
\begin{figure}[htbp]
    \centering
    \includegraphics[scale=0.5]{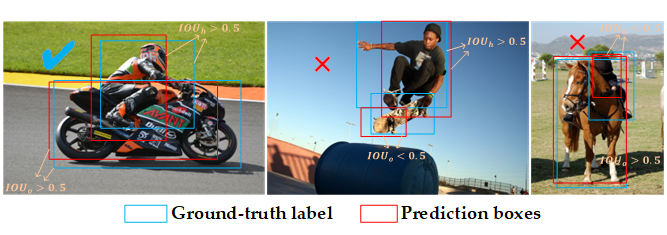}
    \caption{The human-object pairs with human overlap $IOU_h$ and object overlap $IOU_o$ both exceeding 0.5 are declared as true positives. Kindly refer to Sec. \ref{Metric} for more details.}
    \label{fig:Figure9}
\end{figure}

Following the standard evaluation \cite{r47,r59}, we use role mean average precious to evaluate the predicted HOI instances. A detected bounding box is considered a true positive for object detection if it overlaps with a ground truth bounding box of the same class with an intersection greater than union $(IOU)$ greater than 0.5. In HOI detection, we need to predict human-object pairs. The human-object pairs whose human overlap $IOU_h$ and object overlap $IOU_o$ both exceed 0.5, i.e., min $(IOU_h,IOU_o)>0.5$ are declared a true positive (as shown in Fig \ref{fig:Figure9}). Specifically, for HICO-DET, besides the full set of 600 HOI classes, the role mAP over a rare set of 138 HOI classes that have less than 10 training instances and a non-rare set of the other 462 HOI classes are also reported. Furthermore, we report the role mAP of two scenarios for V-COCO: scenario 1 includes the cases even without any objects (for the four action categories of body motions), while scenario 2 ignores these cases. For HOI-SDC, we report the role mean average precision for the full set of 321 HOI classes.

\subsection{Implementation Details}
The Visual Feature Extractor consists of Swin Transformer and a deformable transformer encoder. For Swin-Tiny and Swin-Large, the dimensions of the feature maps in the first stage are set to $C_s=96$ and $C_s=192$, respectively. We pre-train Swin-Tiny on the ImageNet-1k dataset. Swin-Large is first pre-trained on the ImageNet-22k dataset and finetuned on the ImageNet-1k dataset. Then the weights are used to fine-tune the FGAHOI for the HOI detection task. The number of both encoder and decoder layers are set to 6 $(N_{Layer}=6)$. The number of query embeddings is set to 300 $(N_q=300)$, and the hidden dimension of embeddings in the transformer is set to 256 $(C_d=256)$. In the post-processing phase, the first 100 HOI instances are selected according to object confidence, and we use $\delta$=0.5 to filter the HOI instances by the combined $IOU$. Following Deformable-DETR \cite{r15}, the AdamW \cite{r57} optimizer is used. The learning rates of the extractor and the other components are set to $10^{-5}$ and $10^{-4}$, respectively. We use 8 RTX 3090 to train the model (QAHOI $\&$ FGAHOI) with Swin-Tiny. For the model with $\text{Swin-Large}_+^*$, we use 16 RTX 3090 to train them.
For HICO-DET and HOI-SDC, we train the base network for 150 epochs and carry out the learning rate drop from the 120th epoch at the first stage of training. For subsequent training, we trained the model for 40 epochs, with a learning rate drop at the 15th epoch. For V-COCO dataset, we train the base network for 90 epochs and drop the learning rate from 60th epoch at the first stage of training. For subsequent training, we trained the model for 30 epochs, with a learning rate drop at the 10th epoch. 
\begin{figure*}[htbp]
    \centering
    \includegraphics[width = \textwidth]{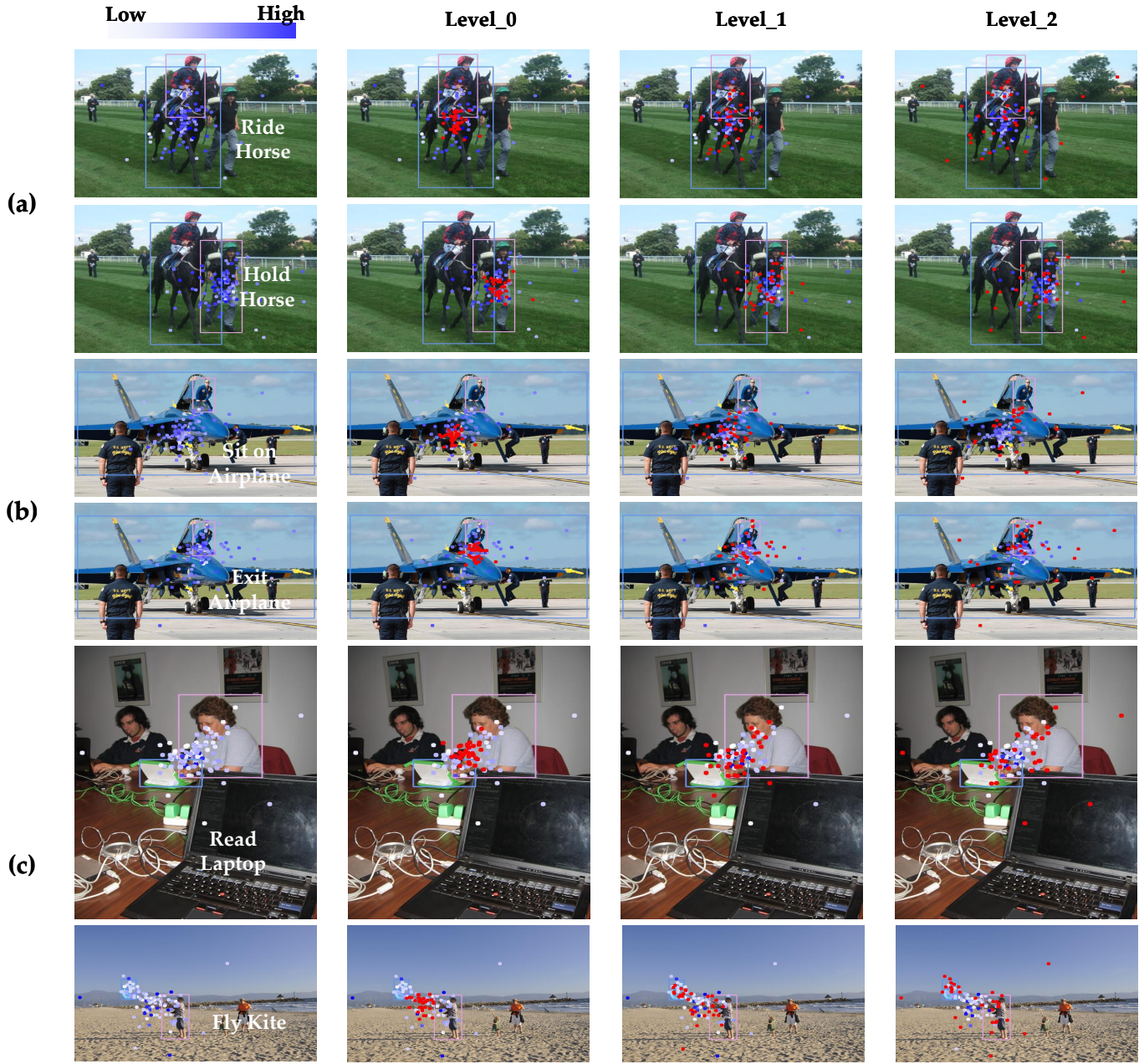}
    \caption{Visualization of fine-grained anchors in the decoding phase, Level\_0, Level\_1 and Level\_2 represent the features at different scales respectively, the color of the blue dots from light to dark represents the degrees of attention of the fine-grained anchors and red dots represent the positions of interest of fine-grained anchors in current scale features. Kindly refer to Sec. \ref{FGA_lookat} for more details.
    % We discover that dynamic anchor is mainly responsible for extracting information near itself in low stage features , while dynamic anchor in high stage features is responsible for exploring distant information. By focusing on different interaction features, dynamic anchor guides the model in identifying different interactive actions for completely different human-object pairs, human-object pairs with the same object and different persons, human-object pairs with different objects and the same human and human-object pairs with the same object and the same human, as shown in (a), (b), (c) and (d), respectively.
    }
    \label{fig:Figure10}
\end{figure*}
\begin{figure*}[htbp]
    \centering
    \includegraphics[width = \textwidth]{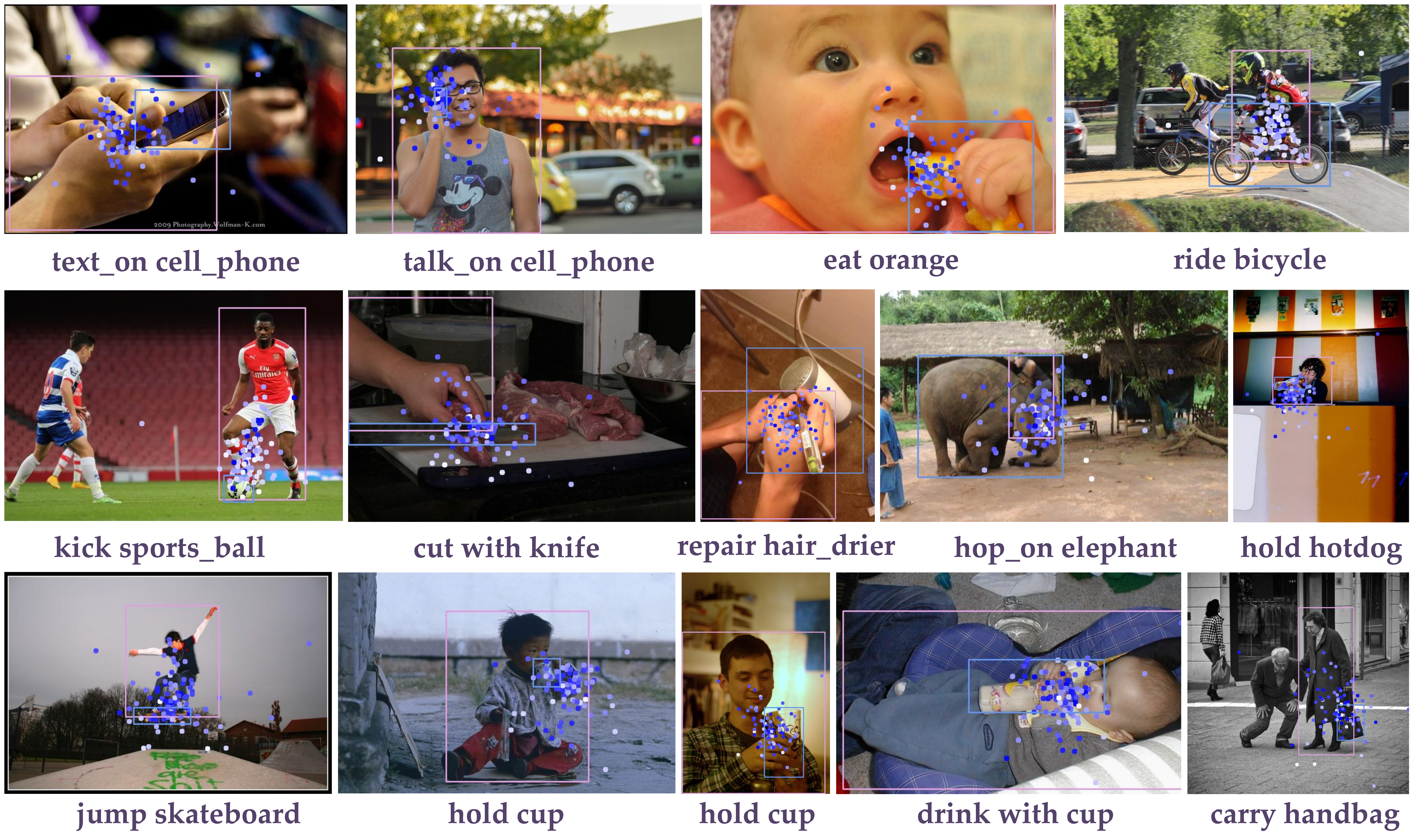}
    \caption{Visualization of several representative interactive actions and the corresponding fine-grained anchors. We only visualize a single representative interactive action for one human-object pair. Kindly refer to Sec. \ref{FGA_lookat} for more details.}
    \label{fig:Figure11}
\end{figure*}

\subsection{Comparison with State-of-the-Arts}
\subsubsection{HICO-DET} \label{sec.sota-hoico-det}
We compare FGAHOI with the state-of-the-art two-stage and one-stage methods on the HICO-DET dataset and report the results in Table.\ref{table 1}. FGAHOI outperforms both state-of-the-art methods. In contrast to the state-of-the-art two-stage method SCG \cite{r24}, FGAHOI with Swin-Large*+ backbone exceeds an especially significant gain of 5.85 mAP in default full setting, 5.99 mAP in default rare setting, 5.8 mAP in default non-rare setting, 4.56 mAP in known object full setting, 4.75 mAP in known rare settings and 4.52 mAP in known object non-rare setting. For a fair comparison, we used the same machine for the reproduction of the QAHOI (as shown in Table.\ref{table 2} QAHOI(R)). In comparison to the state-of-the-art one-stage method QAHOI, FGAHOI exceeds it in all settings for all backbone networks. For Swin-Tiny backbone network, FGAHOI exceeds an especially significant gain of 2.27 mAP in default full setting, 2.02 mAP in default rare setting, 2.55 mAP in default non-rare setting, 2.42 mAP in known object full setting, 1.11 mAP in known rare settings and 2.79 mAP in known object non-rare setting. In addition, FGAHOI with Swin-Large*+ backbone exceeds an especially significant gain of 1.75 mAP in default full setting, 1.49 mAP in default rare setting, 1.82 mAP in default non-rare setting, 1.7 mAP in known object full setting, 0.92 mAP in known rare settings and 1.93 mAP in known object non-rare setting.

\subsubsection{HOI-SDC} \label{HOI-SDC}
On the dataset we propose, $i.e.$, HOI-SDC, we compare FGAHOI with QAHOI and ablate each component of FGAHOI (As shown in Table.\ref{table 4}). The backbone is set to Swin-Tiny. The baseline exceeds QAHOI an especially significant gain of 1.63 mAP. HSAM and TAM improve a significant gain of 0.73 and 0.66 mAP, respectively. Benefit from the MSS, HSAM and TAM, FGAHOI achieve 22.25 mAP on HOI-SDC.

\subsubsection{V-COCO} \label{V-COCO}
We compare FGAHOI with the state-of-the-art methods on V-COCO dataset and report the results in Table.\ref{table 3}. In comparison to QAHOI, FGAHOI only exceeds a small margin. This phenomenon is mainly caused by too little training data in the dataset.
We investigate that FGAHOI cannot adequately perform when the training data is not sufficient due to the complex task requirements. In addition, we investigate the transformer backbone is still superior to CNN backbone in this case.

\subsection{Sensitivity Analysis for UDA and LDVM} \label{Sensitivity Analysis}
% \begin{figure*}[htbp]
%     \centering
%     \includegraphics[width = \textwidth]{figure7 .pdf}
%     \caption{The architecture of FGAHOI's decoder. (a) Illustration of FGAHOI's decoding process. (b) Illustration of Multi-scale sampling strategy. (c) Illustration of Spatial-aware merging mechanism. (d) Illustration of Scale-aware merging mechanism. (e) Illustration of Task-aware merging mechanism. (f) Generation process of fine-grained anchors and the corresponding attention weights.}
% \end{figure*}

According to the two proposed challenges, we divide the HICO-DET into ten intervals. At each intervals, we compare FGAHOI and QAHOI with Swin-Tiny, $\text{Large}^*_+$ backbone, respectively (As shown in Table.\ref{table 5}). When compared between each interval of UDA and LDVM, we investigate that the difficulty of HOI detection decreases as the interval level increases. This justifies the original design. Thus, it is imperative to consider ability of the model to address these two challenges when proposing novel frameworks for HOI detection. In the comparison between FGAHOI and QAHOI, the results demonstrate that FGAHOI has better capability for uneven distributed area and long distance visual modeling of human-object pairs.

% Table 2
\subsection{Qualitative Analysis}

\subsubsection{Visualized Results} \label{visualized_results}
In order to demonstrate our model, several representative HOI predictions are visualized. As shown in Fig.\ref{fig:Figure5}, our model can pinpoint HOI instances from noisy backgrounds and excels at detecting various complicated HOIs, including one object interacting with different humans, one human engaging in multiple interactions with various objects, multiple interactions within a single pair, and multiple humans engaging in various interactions with various objects. In addition, our model is good at long-range visual modelling, withstanding the impacts of hostile environments and small target identification. Fig.\ref{fig:Figure6} (a) illustrates that FGAHOI has excellent long-range visual modelling capabilities and can accurately identify interactions between human-object pairs far from each other. As Fig.\ref{fig:Figure6} (b) shows, our model has outstanding robustness and can effectively resist disruption from harsh environmental factors, including blurring, blocking and glare. Fig.\ref{fig:Figure6} (c) demonstrates the superior capabilities of FGAHOI to identify small HOI instances.\par

\subsubsection{What do the fine-grained anchors look at?} \label{FGA_lookat}
As shown in Fig.\ref{fig:Figure7}, we compare the fine-grained anchors of FGAHOI and QAHOI. First two HOI instances ($i.e$, hold sport ball and ride motorcycles) exhibit that FGAHOI could better focus on humans, objects and the interaction areas rather than noisy backgrounds. The fourth head of FGAHOI still focuses on the HOI instance, while QAHOI focuses on the background. When detecting instance with a long distance between human and object, FGAHOI could focus on the right position, while QAHOI is like a chicken with its head cut off (As shown in the last HOI instance).
\par
To exhibit the effectiveness of the fine-grained anchors for identifying HOI instances and demonstrate the working mechanism of fine-grained anchors, we visualize the fine-grained anchors of the feature maps at different scales in the decoding phase. In Fig.\ref{fig:Figure10} (a), we visualize the instances of two different humans and one object. As shown in Fig.\ref{fig:Figure10} (b), even for exactly the same human-object pair, the areas of focus vary from one interaction to another. In Fig.\ref{fig:Figure10} (c), we show two instances contain short and long distance between humans and objects, respectively. We investigate that the fine-grained anchors of low level feature map focus on small and fine-grained areas. They play a major role in detecting close range and small HOI instances. The fine-grained anchors of high level feature maps focus on large and coarse-grained areas. It is necessary for detecting long distance and large HOI instances.\par
In order to explore what the fine-grained anchors focus on, we visualize several representative actions in Fig.\ref{fig:Figure11}. Visualization shows that fine-grained anchors could concentrate attention precisely on the location where the interactive action is generated. For example, the fine-grained anchors mainly focus on the hand for 'text\_on cell\_phone', the mouth for 'eat orange' and the ear and the mouth for 'talk\_on cell\_phone'. For 'kick sports\_ball', 'jump skateboard' and 'hop\_on elephant', central areas of interest are around legs and feet, while fine-grained anchors primarily focuses on hands for 'carry handbag', 'repair hair\_drier', 'hold cup', 'hold hotdog' and 'cut with kinfe'. 

\subsection{Ablation Study}
In this subsection, a set of experiments are designed to clearly understand the contribution of each of the constituent components of the proposed methodology: \textbf{Merging mechanism}, \textbf{Multi-Scale Sampling Strategy} and \textbf{Stage-wise Training Strategy}. We conducted all experiments on the HICO-DET dataset.

\subsubsection{Ablating FGAHOI Components} \label{Ablating FGAHOI}
To study the contribution of each of the merging mechanisms in FGAHOI, we design careful ablation experiments in Table.\ref{table 6}. To ensure a fair comparison, the sampling sizes are all set to [1, 3, 5]. For the baseline which does not leverages the hierarchical spatial-aware and task-aware merging mechanism, we use the average and direct summation operation to merge the sampled features and connect embeddings. For the results in the table, the middle results denote the role mAP, the results in the top right corner represent the performance improvement compared to QAHOI and the results in the bottom right corner represent the performance improvement compared to the baseline. In comparison to row 1 (QAHOI), row 2 adds the multi-scale sampling strategy. The results demonstrate that adding the sampling strategy improves the ability of the model to detect HOI instances. The row 3 and 4 show that both hierarchical spatial-aware and task-aware merging mechanism make an essential contribution to the success of FGAHOI. The hierarchical spatial-aware merging mechanism, combined with the task-aware merging mechanism performs better together (row 5) than using either of them separately (row 3 and 4). Thus, each component in FGAHOI has a critical role to play in HOI detection.

\subsubsection{Sensitivity Analysis On Multi-Scale Sampling Sizes}
Our multi-scale sampling strategy samples multi-scale features according to the pre-determined sampling sizes. We vary different sampling sizes to conduct the sensitivity analysis for the sampling strategy and report the results in Table.\ref{table 7}. We find that the sampling strategy is relatively stable. Changes in sampling sizes do not have a significant impact on the performance of FGAHOI. However, there is still a slight degradation in the performance of FGAHOI as the sample size increases. We investigate that as the sample size increases, too many background features around the fine-grained anchors are sampled, resulting in contamination of the sampled features and thus the performance of the model suffers. Hence, for validation, we set the sampling sizes to [1, 3, 5] in all our experiments, which is a sweet spot that balances performance.
% Table 4
\begin{table}[htbp]
\centering
\caption{Comparison between different sampling sizes.}
\large
\label{table 7}
\resizebox{\linewidth}{!}{
\begin{tabular}{c c c c c c c c }
\toprule
% \multicolumn{1}{c/}{\multirow{2}{*}{\textbf{Number of intervals}}} &
\multirow{2}{*}{\tabincell{c}{\textbf{Smpling Size}}}&
\multicolumn{3}{c}{\textbf{Default}} &
\multicolumn{3}{c}{\textbf{Known Object}}\\[0.5ex]
\multicolumn{1}{c}{} & 
\multicolumn{1}{c}{\textbf{Full}} & \textbf{Rare} & \textbf{Non-Rare} & \textbf{Full} & \textbf{Rare} &
\multicolumn{1}{c}{\textbf{Non-Rare}}\\
\midrule
% \multirow{3}{*}{\multicolumn{11}{c}{Two-Stage Methods}}\\
% \multirow{4}{*}{}   {1, 1, 1}&	  29.38&   22.79 &	  31.35&	  32.04&	  25.052&	 34.13\\[0.5ex]
\multirow{1}{*}{}   {[ 1, 3, 5 ]}&	  \textbf{29.94}&   22.24 &	 \textbf{ 32.24}&	  \textbf{32.48}&	  24.16&	 \textbf{34.97}\\[0.5ex]
\multirow{1}{*}{}   {[ 3, 5, 7 ]}&	  29.72&	  \textbf{23.03}&	  31.72&	  32.33&	  \textbf{25.67}&	  34.30\\[0.5ex]
\multirow{1}{*}{}   {[ 5, 7, 9 ]}&	  29.65&	  22.64&	  31.74&	  32.55&	  25.64&	  34.62\\[0.5ex]
\bottomrule
\end{tabular}}
\end{table}

\subsubsection{Training Strategies}
As shown in Table.\ref{table 8}, we leverage the stage-wise and end-to-end training strategy to train FGAHOI, respectively. In the end-to-end training strategy, we train FGAHOI for 150 epochs and the learning rate drop is carried out at the 120th epoch. The stage-wise training strategy promotes 5.96 mAP for default full setting, 4.61 for default rare, 6.36 for default non-rare, 6.04 for known object full, 4.65 for known object rare and 6.46 mAP for known object non-rare setting. In comparison to the end-to-end training strategy, we investigate that the stage-wise training strategy reduces the learning difficulty of the FGAHOI and clarify the learning direction of the model by emphasizing it to learn what it needs at each stage.
\begin{table}[htbp]
\centering
\caption{Comparison between Stage-Wise and End-to-End training approach.}
\large
\label{table 8}
\resizebox{\linewidth}{!}{
\begin{tabular}{c c c c c c c c }
\toprule
% \multicolumn{1}{c/}{\multirow{2}{*}{\textbf{Number of intervals}}} &
\multirow{2}{*}{\tabincell{c}{\textbf{Training Strategy}}}&
\multicolumn{3}{c}{\textbf{Default}} &
\multicolumn{3}{c}{\textbf{Known Object}}\\[0.5ex]
\multicolumn{1}{c}{} & 
\multicolumn{1}{c}{\textbf{Full}} & \textbf{Rare} & \textbf{Non-Rare} & \textbf{Full} & \textbf{Rare} &
\multicolumn{1}{c}{\textbf{Non-Rare}}\\
\midrule
% \multirow{3}{*}{\multicolumn{11}{c}{Two-Stage Methods}}\\
% \multirow{4}{*}{}   {1, 1, 1}&	  29.38&   22.79 &	  31.35&	  32.04&	  25.052&	 34.13\\[0.5ex]
\multirow{1}{*}{}   Stage-Wise&	  \textbf{29.94}&   \textbf{22.24} &	 \textbf{ 32.24}&	  \textbf{32.48}&	  \textbf{24.16}&	 \textbf{34.97}\\[0.5ex]
\multirow{1}{*}{}   End-to-End&	  23.98&	  17.63&	  25.88&	  26.44&	 19.51&	  28.51\\[0.5ex]
\bottomrule
\end{tabular}}
\end{table}

\section{Conclusion}
In this paper, we propose a novel transformer-based human-object interaction detector (FGAHOI) which leverages the input features to generate fine-grained anchors for protecting the detection of HOI instances from noisy backgrounds. We propose a novel training strategy where each component of the model is trained sequentially to clarify the training direction at each stage, for maximizing the savings of the training cost. We propose two novel metrics and a novel dataset, $i.e.$, HOI-SDC for the two challenges (Uneven Distributed Area in Human-Object Pairs and Long Distance Visual Modeling of Human-Object Pairs) of detecting HOI instances. Our extensive experiments on three benchmarks: HICO-DET, HOI-SDC and V-COCO, demonstrate the effectiveness of the proposed FGAHOI. Specifically, FGAHOI outperforms all existing state-of-the-art methods by a large margin.

\ifCLASSOPTIONcompsoc
  \section*{Acknowledgments}
\else
  \section*{Acknowledgment}
\fi

This work is supported by National Natural Science Foundation of China (grant No.61871106 and No.61370152), Key R\&D projects of Liaoning Province, China (grant No.2020JH2/10100029), and the Open Project Program Foundation of the Key Laboratory of Opto-Electronics Information Processing, Chinese Academy of Sciences (OEIP-O-202002).

\ifCLASSOPTIONcaptionsoff
  \newpage
\fi

\bibliographystyle{ieeetr} 
\bibliography{reference}

\begin{thebibliography}{10}

\bibitem{r16}
R.~Girshick, ``Fast r-cnn,'' in {\em Proceedings of the IEEE international
  conference on computer vision}, pp.~1440--1448, 2015.

\bibitem{r22}
Z.~Li and F.~Zhou, ``Fssd: feature fusion single shot multibox detector,'' {\em
  arXiv preprint arXiv:1712.00960}, 2017.

\bibitem{r31}
S.~Ren, K.~He, R.~Girshick, and J.~Sun, ``Faster r-cnn: Towards real-time
  object detection with region proposal networks,'' {\em Advances in neural
  information processing systems}, vol.~28, 2015.

\bibitem{r40}
R.~Girshick, J.~Donahue, T.~Darrell, and J.~Malik, ``Rich feature hierarchies
  for accurate object detection and semantic segmentation,'' in {\em
  Proceedings of the IEEE conference on computer vision and pattern
  recognition}, pp.~580--587, 2014.

\bibitem{r46}
J.~Redmon, S.~Divvala, R.~Girshick, and A.~Farhadi, ``You only look once:
  Unified, real-time object detection,'' in {\em Proceedings of the IEEE
  conference on computer vision and pattern recognition}, pp.~779--788, 2016.

\bibitem{r2}
M.~Chen, Y.~Liao, S.~Liu, Z.~Chen, F.~Wang, and C.~Qian, ``Reformulating hoi
  detection as adaptive set prediction,'' in {\em Proceedings of the IEEE/CVF
  Conference on Computer Vision and Pattern Recognition}, pp.~9004--9013, 2021.

\bibitem{r4}
Z.~Hou, X.~Peng, Y.~Qiao, and D.~Tao, ``Visual compositional learning for
  human-object interaction detection,'' in {\em European Conference on Computer
  Vision}, pp.~584--600, Springer, 2020.

\bibitem{r5}
P.~Zhou and M.~Chi, ``Relation parsing neural network for human-object
  interaction detection,'' in {\em Proceedings of the IEEE/CVF International
  Conference on Computer Vision}, pp.~843--851, 2019.

\bibitem{r7}
B.~Kim, J.~Lee, J.~Kang, E.-S. Kim, and H.~J. Kim, ``Hotr: End-to-end
  human-object interaction detection with transformers,'' in {\em Proceedings
  of the IEEE/CVF Conference on Computer Vision and Pattern Recognition},
  pp.~74--83, 2021.

\bibitem{r8}
X.~Zhong, C.~Ding, X.~Qu, and D.~Tao, ``Polysemy deciphering network for robust
  human--object interaction detection,'' {\em International Journal of Computer
  Vision}, vol.~129, no.~6, pp.~1910--1929, 2021.

\bibitem{r9}
X.~Zhong, X.~Qu, C.~Ding, and D.~Tao, ``Glance and gaze: Inferring action-aware
  points for one-stage human-object interaction detection,'' in {\em
  Proceedings of the IEEE/CVF Conference on Computer Vision and Pattern
  Recognition}, pp.~13234--13243, 2021.

\bibitem{r10}
C.~Gao, J.~Xu, Y.~Zou, and J.-B. Huang, ``Drg: Dual relation graph for
  human-object interaction detection,'' in {\em European Conference on Computer
  Vision}, pp.~696--712, Springer, 2020.

\bibitem{r14}
O.~Ulutan, A.~Iftekhar, and B.~S. Manjunath, ``Vsgnet: Spatial attention
  network for detecting human object interactions using graph convolutions,''
  in {\em Proceedings of the IEEE/CVF conference on computer vision and pattern
  recognition}, pp.~13617--13626, 2020.

\bibitem{r17}
C.~Gao, Y.~Zou, and J.-B. Huang, ``ican: Instance-centric attention network for
  human-object interaction detection,'' {\em arXiv preprint arXiv:1808.10437},
  2018.

\bibitem{r19}
T.~Wang, T.~Yang, M.~Danelljan, F.~S. Khan, X.~Zhang, and J.~Sun, ``Learning
  human-object interaction detection using interaction points,'' in {\em
  Proceedings of the IEEE/CVF Conference on Computer Vision and Pattern
  Recognition}, pp.~4116--4125, 2020.

\bibitem{r20}
A.~Zhang, Y.~Liao, S.~Liu, M.~Lu, Y.~Wang, C.~Gao, and X.~Li, ``Mining the
  benefits of two-stage and one-stage hoi detection,'' {\em Advances in Neural
  Information Processing Systems}, vol.~34, pp.~17209--17220, 2021.

\bibitem{r21}
C.~Zou, B.~Wang, Y.~Hu, J.~Liu, Q.~Wu, Y.~Zhao, B.~Li, C.~Zhang, C.~Zhang,
  Y.~Wei, {\em et~al.}, ``End-to-end human object interaction detection with
  hoi transformer,'' in {\em Proceedings of the IEEE/CVF conference on computer
  vision and pattern recognition}, pp.~11825--11834, 2021.

\bibitem{r24}
F.~Z. Zhang, D.~Campbell, and S.~Gould, ``Spatially conditioned graphs for
  detecting human-object interactions,'' in {\em Proceedings of the IEEE/CVF
  International Conference on Computer Vision}, pp.~13319--13327, 2021.

\bibitem{r33}
M.~Tamura, H.~Ohashi, and T.~Yoshinaga, ``Qpic: Query-based pairwise
  human-object interaction detection with image-wide contextual information,''
  in {\em Proceedings of the IEEE/CVF Conference on Computer Vision and Pattern
  Recognition}, pp.~10410--10419, 2021.

\bibitem{r39}
Y.-L. Li, S.~Zhou, X.~Huang, L.~Xu, Z.~Ma, H.-S. Fang, Y.~Wang, and C.~Lu,
  ``Transferable interactiveness knowledge for human-object interaction
  detection,'' in {\em Proceedings of the IEEE/CVF Conference on Computer
  Vision and Pattern Recognition}, pp.~3585--3594, 2019.

\bibitem{r47}
G.~Gkioxari, R.~Girshick, P.~Doll{\'a}r, and K.~He, ``Detecting and recognizing
  human-object interactions,'' in {\em Proceedings of the IEEE conference on
  computer vision and pattern recognition}, pp.~8359--8367, 2018.

\bibitem{r30}
Y.-W. Chao, Y.~Liu, X.~Liu, H.~Zeng, and J.~Deng, ``Learning to detect
  human-object interactions,'' in {\em 2018 ieee winter conference on
  applications of computer vision (wacv)}, pp.~381--389, IEEE, 2018.

\bibitem{r38}
T.~Gupta, A.~Schwing, and D.~Hoiem, ``No-frills human-object interaction
  detection: Factorization, layout encodings, and training techniques,'' in
  {\em Proceedings of the IEEE/CVF International Conference on Computer
  Vision}, pp.~9677--9685, 2019.

\bibitem{r43}
B.~Wan, D.~Zhou, Y.~Liu, R.~Li, and X.~He, ``Pose-aware multi-level feature
  network for human object interaction detection,'' in {\em Proceedings of the
  IEEE/CVF International Conference on Computer Vision}, pp.~9469--9478, 2019.

\bibitem{r44}
A.~Bansal, S.~S. Rambhatla, A.~Shrivastava, and R.~Chellappa, ``Detecting
  human-object interactions via functional generalization,'' in {\em
  Proceedings of the AAAI Conference on Artificial Intelligence}, vol.~34,
  pp.~10460--10469, 2020.

\bibitem{r12}
H.-S. Fang, Y.~Xu, W.~Wang, X.~Liu, and S.-C. Zhu, ``Learning pose grammar to
  encode human body configuration for 3d pose estimation,'' in {\em Proceedings
  of the AAAI conference on artificial intelligence}, vol.~32, 2018.

\bibitem{r32}
H.-S. Fang, G.~Lu, X.~Fang, J.~Xie, Y.-W. Tai, and C.~Lu, ``Weakly and semi
  supervised human body part parsing via pose-guided knowledge transfer,'' {\em
  arXiv preprint arXiv:1805.04310}, 2018.

\bibitem{r37}
Y.~Xiu, J.~Li, H.~Wang, Y.~Fang, and C.~Lu, ``Pose flow: Efficient online pose
  tracking,'' {\em arXiv preprint arXiv:1802.00977}, 2018.

\bibitem{r29}
S.~Qi, W.~Wang, B.~Jia, J.~Shen, and S.-C. Zhu, ``Learning human-object
  interactions by graph parsing neural networks,'' in {\em Proceedings of the
  European conference on computer vision (ECCV)}, pp.~401--417, 2018.

\bibitem{r41}
B.~Kim, T.~Choi, J.~Kang, and H.~J. Kim, ``Uniondet: Union-level detector
  towards real-time human-object interaction detection,'' in {\em European
  Conference on Computer Vision}, pp.~498--514, Springer, 2020.

\bibitem{r45}
Y.~Liao, S.~Liu, F.~Wang, Y.~Chen, C.~Qian, and J.~Feng, ``Ppdm: Parallel point
  detection and matching for real-time human-object interaction detection,'' in
  {\em Proceedings of the IEEE/CVF Conference on Computer Vision and Pattern
  Recognition}, pp.~482--490, 2020.

\bibitem{r3}
N.~Carion, F.~Massa, G.~Synnaeve, N.~Usunier, A.~Kirillov, and S.~Zagoruyko,
  ``End-to-end object detection with transformers,'' in {\em European
  conference on computer vision}, pp.~213--229, Springer, 2020.

\bibitem{r34}
J.~Chen and K.~Yanai, ``Qahoi: Query-based anchors for human-object interaction
  detection,'' {\em arXiv preprint arXiv:2112.08647}, 2021.

\bibitem{r15}
X.~Zhu, W.~Su, L.~Lu, B.~Li, X.~Wang, and J.~Dai, ``Deformable detr: Deformable
  transformers for end-to-end object detection,'' {\em arXiv preprint
  arXiv:2010.04159}, 2020.

\bibitem{r73}
Y.~Bengio, P.~Lamblin, D.~Popovici, and H.~Larochelle, ``Greedy layer-wise
  training of deep networks,'' {\em Advances in neural information processing
  systems}, vol.~19, 2006.

\bibitem{r74}
G.~E. Hinton, S.~Osindero, and Y.-W. Teh, ``A fast learning algorithm for deep
  belief nets,'' {\em Neural computation}, vol.~18, no.~7, pp.~1527--1554,
  2006.

\bibitem{r75}
B.~Kang, S.~Xie, M.~Rohrbach, Z.~Yan, A.~Gordo, J.~Feng, and Y.~Kalantidis,
  ``Decoupling representation and classifier for long-tailed recognition,''
  {\em arXiv preprint arXiv:1910.09217}, 2019.

\bibitem{r54}
Y.-W. Chao, Y.~Liu, X.~Liu, H.~Zeng, and J.~Deng, ``Learning to detect
  human-object interactions,'' in {\em 2018 ieee winter conference on
  applications of computer vision (wacv)}, pp.~381--389, IEEE, 2018.

\bibitem{r59}
S.~Gupta and J.~Malik, ``Visual semantic role labeling,'' {\em arXiv e-prints},
  2015.

\bibitem{r48}
F.~Scarselli, M.~Gori, A.~C. Tsoi, M.~Hagenbuchner, and G.~Monfardini, ``The
  graph neural network model,'' {\em IEEE transactions on neural networks},
  vol.~20, no.~1, pp.~61--80, 2008.

\bibitem{r11}
A.~Vaswani, N.~Shazeer, N.~Parmar, J.~Uszkoreit, L.~Jones, A.~N. Gomez,
  {\L}.~Kaiser, and I.~Polosukhin, ``Attention is all you need,'' {\em Advances
  in neural information processing systems}, vol.~30, 2017.

\bibitem{r49}
Y.~LeCun, Y.~Bengio, {\em et~al.}, ``Convolutional networks for images, speech,
  and time series,'' {\em The handbook of brain theory and neural networks},
  vol.~3361, no.~10, p.~1995, 1995.

\bibitem{r27}
Z.~Liu, Y.~Lin, Y.~Cao, H.~Hu, Y.~Wei, Z.~Zhang, S.~Lin, and B.~Guo, ``Swin
  transformer: Hierarchical vision transformer using shifted windows,'' in {\em
  Proceedings of the IEEE/CVF International Conference on Computer Vision},
  pp.~10012--10022, 2021.

\bibitem{r63}
X.~Chen, F.~Wei, G.~Zeng, and J.~Wang, ``Conditional detr v2: Efficient
  detection transformer with box queries,'' {\em arXiv preprint
  arXiv:2207.08914}, 2022.

\bibitem{r60}
Y.~Wang, X.~Zhang, T.~Yang, and J.~Sun, ``Anchor detr: Query design for
  transformer-based detector,'' in {\em Proceedings of the AAAI conference on
  artificial intelligence}, vol.~36, pp.~2567--2575, 2022.

\bibitem{r61}
S.~Liu, F.~Li, H.~Zhang, X.~Yang, X.~Qi, H.~Su, J.~Zhu, and L.~Zhang,
  ``Dab-detr: Dynamic anchor boxes are better queries for detr,'' {\em arXiv
  preprint arXiv:2201.12329}, 2022.

\bibitem{r64}
G.~Zhang, Z.~Luo, Y.~Yu, K.~Cui, and S.~Lu, ``Accelerating detr convergence via
  semantic-aligned matching,'' in {\em Proceedings of the IEEE/CVF Conference
  on Computer Vision and Pattern Recognition}, pp.~949--958, 2022.

\bibitem{r70}
K.~He, X.~Zhang, S.~Ren, and J.~Sun, ``Deep residual learning for image
  recognition,'' in {\em Proceedings of the IEEE conference on computer vision
  and pattern recognition}, pp.~770--778, 2016.

\bibitem{r71}
F.~Iandola, M.~Moskewicz, S.~Karayev, R.~Girshick, T.~Darrell, and K.~Keutzer,
  ``Densenet: Implementing efficient convnet descriptor pyramids,'' {\em arXiv
  preprint arXiv:1404.1869}, 2014.

\bibitem{r72}
A.~Newell, K.~Yang, and J.~Deng, ``Stacked hourglass networks for human pose
  estimation,'' in {\em European conference on computer vision}, pp.~483--499,
  Springer, 2016.

\bibitem{r1}
X.~Dong, J.~Bao, D.~Chen, W.~Zhang, N.~Yu, L.~Yuan, D.~Chen, and B.~Guo,
  ``Cswin transformer: A general vision transformer backbone with cross-shaped
  windows,'' in {\em Proceedings of the IEEE/CVF Conference on Computer Vision
  and Pattern Recognition}, pp.~12124--12134, 2022.

\bibitem{r13}
J.~Yang, C.~Li, P.~Zhang, X.~Dai, B.~Xiao, L.~Yuan, and J.~Gao, ``Focal
  self-attention for local-global interactions in vision transformers,'' {\em
  arXiv preprint arXiv:2107.00641}, 2021.

\bibitem{r23}
Z.~Xia, X.~Pan, S.~Song, L.~E. Li, and G.~Huang, ``Vision transformer with
  deformable attention,'' in {\em Proceedings of the IEEE/CVF Conference on
  Computer Vision and Pattern Recognition}, pp.~4794--4803, 2022.

\bibitem{r26}
H.~Wu, B.~Xiao, N.~Codella, M.~Liu, X.~Dai, L.~Yuan, and L.~Zhang, ``Cvt:
  Introducing convolutions to vision transformers,'' in {\em Proceedings of the
  IEEE/CVF International Conference on Computer Vision}, pp.~22--31, 2021.

\bibitem{r35}
A.~Dosovitskiy, L.~Beyer, A.~Kolesnikov, D.~Weissenborn, X.~Zhai,
  T.~Unterthiner, M.~Dehghani, M.~Minderer, G.~Heigold, S.~Gelly, {\em et~al.},
  ``An image is worth 16x16 words: Transformers for image recognition at
  scale,'' {\em arXiv preprint arXiv:2010.11929}, 2020.

\bibitem{r36}
C.-F. Chen, R.~Panda, and Q.~Fan, ``Regionvit: Regional-to-local attention for
  vision transformers,'' {\em arXiv preprint arXiv:2106.02689}, 2021.

\bibitem{r42}
Y.~Wang, R.~Huang, S.~Song, Z.~Huang, and G.~Huang, ``Not all images are worth
  16x16 words: Dynamic transformers for efficient image recognition,'' {\em
  Advances in Neural Information Processing Systems}, vol.~34,
  pp.~11960--11973, 2021.

\bibitem{r68}
K.~He, G.~Gkioxari, P.~Doll{\'a}r, and R.~Girshick, ``Mask r-cnn,'' in {\em
  Proceedings of the IEEE international conference on computer vision},
  pp.~2961--2969, 2017.

\bibitem{r69}
X.~Dai, Y.~Chen, B.~Xiao, D.~Chen, M.~Liu, L.~Yuan, and L.~Zhang, ``Dynamic
  head: Unifying object detection heads with attentions,'' in {\em Proceedings
  of the IEEE/CVF conference on computer vision and pattern recognition},
  pp.~7373--7382, 2021.

\bibitem{r66}
H.~Law and J.~Deng, ``Cornernet: Detecting objects as paired keypoints,'' in
  {\em Proceedings of the European conference on computer vision (ECCV)},
  pp.~734--750, 2018.

\bibitem{r65}
T.-Y. Lin, P.~Goyal, R.~Girshick, K.~He, and P.~Doll{\'a}r, ``Focal loss for
  dense object detection,'' in {\em Proceedings of the IEEE international
  conference on computer vision}, pp.~2980--2988, 2017.

\bibitem{r67}
Y.-L. Li, L.~Xu, X.~Liu, X.~Huang, Y.~Xu, S.~Wang, H.-S. Fang, Z.~Ma, M.~Chen,
  and C.~Lu, ``Pastanet: Toward human activity knowledge engine,'' in {\em
  Proceedings of the IEEE/CVF Conference on Computer Vision and Pattern
  Recognition}, pp.~382--391, 2020.

\bibitem{r56}
Y.-W. Chao, Z.~Wang, Y.~He, J.~Wang, and J.~Deng, ``Hico: A benchmark for
  recognizing human-object interactions in images,'' in {\em Proceedings of the
  IEEE international conference on computer vision}, pp.~1017--1025, 2015.

\bibitem{r57}
I.~Loshchilov and F.~Hutter, ``Decoupled weight decay regularization,'' {\em
  arXiv preprint arXiv:1711.05101}, 2017.

\end{thebibliography}

\end{document}